\pdfoutput=1

\documentclass[11pt]{article}

\usepackage[]{acl}

\usepackage{times}
\usepackage{latexsym}
\usepackage{enumitem}
\usepackage{afterpage}
\usepackage{capt-of}
\usepackage{comment}
\usepackage[T1]{fontenc}
\usepackage{amssymb}
\usepackage{pifont}
\newcommand{\cmark}{\ding{51}}%
\newcommand{\xmark}{\ding{55}}%

\usepackage[utf8]{inputenc}
\usepackage[greek,english]{babel}

\usepackage{microtype}
\usepackage{longtable}
\usepackage{hyperref}

\usepackage{inconsolata}
\usepackage{booktabs}
\usepackage{multirow}
\usepackage{graphicx}
\usepackage{textgreek}

\usepackage[normalem]{ulem}

\newcommand{\dataset}[0]{\textsc{CoDET}}

\title{\dataset: A Benchmark for \textit{Co}ntrastive \textit{D}ialectal \textit{E}valuation \\of Machine \textit{T}ranslation}

\author{Md Mahfuz Ibn Alam$^\alpha$ \qquad Sina Ahmadi$^{\alpha,\beta}$ \qquad Antonios Anastasopoulos$^{\alpha,\gamma}$ \\
        $^\alpha$Department of Computer Science, George Mason University \hfill $^{\beta}$University of Zurich \\ 
        $^\gamma$Archimedes AI Research Unit, RC Athena, Greece\\
        \texttt{\{malam21,sahmad46,antonis\}@gmu.edu}}

\begin{document}
\maketitle
\begin{abstract}
Neural machine translation (NMT) systems exhibit limited robustness in handling source-side linguistic variations. Their performance tends to degrade when faced with even slight deviations in language usage, such as different domains or variations introduced by second-language speakers. It is intuitive to extend this observation to encompass dialectal variations as well, but the work allowing the community to evaluate MT systems on this dimension is limited. To alleviate this issue, we compile and release \dataset, a contrastive dialectal benchmark encompassing 891 different variations from twelve different languages. We also quantitatively demonstrate the challenges large MT models face in effectively translating dialectal variants. All the data and code\footnote{\url{https://github.com/mahfuzibnalam/dialect_mt}} have been released.
\end{abstract}

\section{Introduction}
Progress in natural language processing (NLP) and other varieties of human language technology throughout the 2010s has been undeniably swift. However, such advances are limited to a set of languages with largely available resources ~\cite{joshi-etal-2020-state,blasi-etal-2022-systematic}; they have focused solely on dominant, "standard" language varieties. But no language is a monolith; languages vary richly across countries, regions, social classes, and other factors\footnote{In this paper, we will use the terms ``dialect'' and ``language variety'' interchangeably for readability reasons. The distinction between what is named a language and what a dialect or variety is a complex socioeconomic phenomenon rather than a purely linguistic one. We add a bit of discussion in Section~\ref{sec:dataset} for each variety/language we work with.}.

For modern \textit{linguae francae} such as English, Spanish, or French, some commercial systems apply coarse localization, e.g., Google Assistant supports speech recognition for English in at least seven locales.\footnote{(AU, CA, GB, IN, BE, SG, US)} This, however, is not the case for the majority of the world's languages, even if they exhibit large variations across dialects and regions, often corresponding to millions of speakers. As a result, we have a limited understanding of how well modern NLP systems can handle (or not) such data. It is crucial that we first quantify such disparities in as many languages as possible before we explore ways of mitigating any performance imbalances we identify.

Language variants can vary along several dimensions. In this work, we focus on the robust \textit{understanding} of lexical and morphosyntactic variations, which show up in the written form of languages and hence can be evaluated through a downstream task like text-based machine translation. If one wanted to capture phonological variation additionally, one should work directly on audio and tasks like automatic speech recognition or speech translation; we leave this vein of work for the future.
\begin{table}[!t]
    \centering
    \begin{tabular}{r|cc}
    \toprule
        \multicolumn{3}{l}{Standard Italian Variant:} \\
        Source: & \textit{Hanno rubato il quadro} \\
        GTranslate: & They stole the painting & \cmark\\ \midrule
        \multicolumn{3}{l}{Alassio Variant:} \\ 
        Source: & \textit{I han rubbau u quaddru}\\
        GTranslate: & I han rubbau u quaddru &  \xmark\\
    \bottomrule
    \end{tabular}
    \vspace*{-2mm}
    \caption{While it properly translates standard Italian into English, a popular translation system utterly fails to translate the Alassio variety. \textit{Contrastive dialectal} examples like this one, even if short, can reveal and properly quantify such inadequacies in MT performance.}
    \label{tab:difference}
    \vspace*{-6mm}
\end{table}

Consider the case study presented in Table~\ref{tab:difference}: given two sentences that have the same meaning,\footnote{Correct translation: \textit{``They stole the painting''.}} Google Translate produces very different results. In the first, in "standard" Italian, it produces a perfect translation. The second, from the variety spoken in Alassio in Northwest Italy, the MT system fails to produce any English translation, simply copying the source. Our assumption for evaluating the system is that both inputs should yield the same translated output. This example effectively illustrates the limitations of general MT systems in comprehending and accurately translating dialectal variations.

To properly evaluate such inadequacies in the context of machine translation, one needs \textit{contrastive} examples between varieties so that the evaluation metrics are comparable. Our work attempts to fill this gap.
In summary, our contributions are: 
\begin{itemize}[noitemsep,nolistsep,leftmargin=*]
    \item We extract contrastive data from previous dialectology studies in three languages: Italian (439 locales), Basque (39 locales), and Swiss German (368 locales);
    \item We re-purpose contrastive data from various sources in seven languages: Arabic (25 vernaculars), Occitan (2 varieties), Tigrinya (2 varieties), Farsi (2 varieties), Malay-Indonesian (2 varieties), Swahili (2 varieties), and Greek (1 variety);
    \item We create a limited amount of contrastive data in additional languages: Bengali (5 varieties) and Central Kurdish (4 varieties).
    \item We benchmark the selected distinct dialects of the target language using state-of-the-art machine translation models and quantify the performance discrepancies across language varieties.
\end{itemize}

\section{Related Work}
\label{app:related}
MT is one of the most studied and pioneering tasks in the NLP realm. Many previous studies have focused on proposing more efficient methods, particularly with recent advances in sequence-to-sequence models ~\cite{sutskever2014sequence}, attention mechanism ~\cite{bahdanau2014neural}, and transformers ~\cite{vaswani2017attention} that have left their impact on other tasks in NLP as well. Although creating MT models for languages around the globe has received much attention, as in FLORES-200 benchmark and No Language Left Behind (NLLB) models ~\cite{costa2022no}, we have a considerable stretch remaining to create models that can translate dialects and varieties efficiently.

Most of the previous work on developing MT technologies for dialects and varieties address Arabic ~\cite{zbib2012machine,harrat2019machine}, Swiss German ~\cite{garner14_interspeech,honnet2017machine}, Kurdish ~\cite{ahmadi2022leveraging}, Portuguese ~\cite{fancellu2014standard} and French ~\cite{garcia2022using}. In this regard, one of the main challenges is finding possible translation sources and creating corpora and datasets for the translation of varieties and dialects ~\cite{zampieri2020natural}. In the same vein, exploring the translation of varieties in a few-short or zero-shot setting has received attention ~\cite{riley2022frmt}. Similarly, fine-tuning translation models trained on closely related languages has been proposed as a remedy ~\cite{kumar-etal-2021-machine}.

Given that there is currently no benchmark for the existing data on MT of dialects and varieties, our paper aims to provide one with the sole objective of evaluating varieties and the performance and resilience of MT models to dialectal variations. We also believe this work will increase awareness of this task and motivate future efforts.

\section{The \dataset{} Benchmark}
\label{sec:dataset}

Given a sentence in one dialectal variant and another in the standard variant of the same language as in Table~\ref{tab:difference}, if these two sentences have the same meaning, we can call this \emph{contrastive} of each other. While these data are also \textit{parallel}, we prefer to point to the contrast between the two, as is common in the comparative dialectology literature. The term "parallel" has been widely used to refer to the interlingual aspect of translation, so we wanted to avoid confusion.

Given that little has been done in this vein, we focus on creating constructive datasets following three approaches, namely repurposing previous dialectological work on syntactic variations for Basque, Italian, Swiss German, and Central Occita; manual translation by native dialect speakers for Bengali, Modern Greek, Central Kurdish; and finally, exploiting some existing resources for Arabic, Farsi, Malay-Indonesian, Tigrinya, and Swahili. Table~\ref{tab:sentences} provides the number of sentences along with the number of varieties that the dataset covers.

\begin{table}[!t]
    \centering
    \begin{tabular}{l@{ }@{ }c@{ }@{ }c@{}}
    \toprule
    \multirow{1}{*}{\textbf{Languages/Varieties}} & \multirow{1}{*}{\textbf{\# Sents}} & \multirow{1}{*}{\textbf{\# Varieties}}\\ \midrule
    Italian Varieties & 792 & 439\\
    Swiss German Varieties& 118 & 368\\
    Basque Varieties & 370 & 39\\
    Arabic Vernaculars & 12,000 & 25\\
    Bengali Varieties & 200 & 5\\
    Central Kurdish Varieties & 300 & 4\\
    Farsi Varieties & 3071 & 2\\
    Malay-Indonesian & 3071 & 2\\
    Swahili & 1919 & 2\\
    Tigrinya Varieties & 3071 & 2\\
    Aranese & 476 & 1\\
    Central Occitan & 379 & 1\\
    Griko & 163 & 1\\
    \bottomrule
    \end{tabular}
    \vspace*{-2mm}
    \caption{Number of contrastive sentences in \dataset.}
    \label{tab:sentences}
    \vspace*{-6mm}
\end{table}
\vspace*{-2mm}
\paragraph{Utilizing Existing Datasets} A small amount of work has already provided contrastive examples for varieties of some languages. Some were created as part of dialectological work, which we manually scraped from dissertations and theses; some were created as part of other efforts, such as the TICO-19 and the MADAR corpora.\footnote{See details below.} 

\vspace*{-2mm}
\paragraph{Scraping Syntactic Atlases}
Traditionally, researchers and fieldworkers employ questionnaires to individuals fluent in specific dialects to gather the necessary data for dialectological studies. The questionnaires are designed to elicit responses regarding how a particular sentence or phrase would be expressed in their respective dialects, as in ``how do you say this sentence... in your dialect?'' where the speaker fills the gap based on the target dialect.\footnote{An alternative approach pre-constructs sentence examples and elicits grammatical responses from the informants.} This systematic approach allows for the collection of dialectal data that serves as a valuable resource for investigating the linguistic changes in different varieties and for comprehensively examining and analyzing the variations between the dialects.

Although describing and documenting dialectal variations in most languages have received limited attention in the research landscape, notable efforts\footnote{We talk about these efforts in Section\ref{sec:languages}} have been made to study variations in some European languages, such as Italian, Basque, and Swiss German, through the creation of syntactic atlases.
\vspace*{-2mm}
\paragraph{New Data Creation} For a handful of languages, namely Central Kurdish, Bengali, Griko, and Occitan, we did not find any existing dialectal contrastive data, but we were able to construct small evaluation benchmarks by online data scraping (Occitan) and by reaching out to native speakers and translators of these varieties (for the others).

\subsection{The Languages of \dataset} \label{sec:languages}
We direct the interested reader to Appendix~\ref{app:languages}, where we discuss each of the languages/varieties included in our benchmark.
Due to space limitations, below we only briefly list the languages and varieties included in \dataset.

First, the data sourced from Syntactic Atlases:
\begin{itemize}[noitemsep,nolistsep,leftmargin=*]
    \item \textbf{Basque Varieties:} Our Basque data is sourced from the Basque Syntactic Database.\footnote{\url{http://ixa2.si.ehu.eus/atlas2/index.php}} The data are $n$-way parallel between 39 varieties of the Northern Basque Country in France and come with translations in French and English.
    \item \textbf{Italian Varieties and Languages:} We obtain data from the Italian Syntactic Atlas\footnote{\url{http://svrims2.dei.unipd.it:8080/asit-maldura/pages/search.jsp}} which provides a rich collection of 439 varieties from almost all regions of Italy. We note that many vernaculars spoken around Italy are recognized as officially distinct languages (e.g., Neapolitan, Ligurian, and Venetian, to name a few). Some of these also have a distinct online presence (e.g., with decent Wikipedias), and some MT research is devoted to them ~\cite{nllbteam2022language}. However, this "discretization" of the language continuum observed in the Italian peninsula, where each city/village is said to have its dialect, is far from realistic.
    \item \textbf{Swiss German Varieties:} We obtain data by scraping the Syntactic Atlas of German Switzerland (SADS).\footnote{\url{https://dialektsyntax.linguistik.uzh.ch}} The SADS website hosts a total of 118 questionnaires, each accompanied by answers provided in 368 different locales, all $n$-way parallel along with standard Swiss German.
\end{itemize}

Second, we repurpose an existing dataset:
\begin{itemize}[noitemsep,nolistsep]
    \item \textbf{Arabic Vernaculars:} While Modern Standard Arabic (MSA) is the standardized form of the language used across various regions, MSA is not the native language of Arabic speakers. In informal and spontaneous settings where spoken MSA is typically expected, such as in TV talk shows, speakers often code-switch between their respective vernaculars and MSA.
    To examine MT performance in Arabic dialects, we repurpose the MADAR corpus ~\cite{bouamor-etal-2018-madar}, which consists of 12,000 sentences on varieties from 25 different Arabic-speaking cities, 2,000 of which are $n$-way parallel.
\end{itemize}

Third, we include data from existing MT benchmarks that encompass dialectal variations. In particular, we include some languages from the TICO-19 dataset~\cite{anastasopoulos-etal-2020-tico}, which provides professionally created translations of the same 3071 English sentences related to the COVID-19 domain. We use the following language varieties (all of which are parallel):
\begin{itemize}[noitemsep,nolistsep,leftmargin=*]
    \item \textbf{Tigrinya:} Translations localized to both Ethiopia and Eritrea.
    \item \textbf{Farsi and Dari:} We have translations into Farsi as spoken in Iran and Dari, one of the Farsi variants spoken in Afghanistan.
    \item \textbf{Malay and Indonesian:} We have data in Malay and one of its standardized variants, Indonesian.
    \item \textbf{Swahili:} The TICO-19 dataset provides Coastal Swahili translations (as spoken in Kenya/Tanzania). A follow-up project also provided Congolese Swahili ones~\cite{anastasopoulos-etal-2021-findings}.
\end{itemize}

\noindent Last, we curate new datasets:
\begin{itemize}[noitemsep,nolistsep,leftmargin=*]
    \item \textbf{Bengali Varieties:} Anecdotally, Bangladesh witnesses a linguistic transition approximately every 10 miles. This work specifically focuses on five prominent dialects from five locales of Bangladesh: Jessore, Khulna, Kushtia, Barisal, and Dhaka. The selection of these dialects was strategic, encompassing regions both close to the origin of standard Bengali (Jessore, Kushtia) and those situated farther away.
    
    Our approach involved initially gathering 200 standard Bengali sentences from the Bengali-English translation dataset presented in ~\cite{hasan-etal-2020-low}, a high-quality dataset comprising 2.75 million parallel sentence pairs. From this dataset, we selected short sentences comprising 6 to 7 words, facilitating ease of translation for the language speakers. Initially, there were 200,000 sentences to choose from, and we randomly selected 200 sentences for our dataset.
    
    Our initial step involved recruiting proficient annotators fluent in the standard and in one of the dialects. Subsequently, we requested these annotators to provide their respective dialectal renditions of specific sentences. Given that dialects primarily exist in spoken form without standardized orthography, we instructed the annotators to transcribe the sentences in Bengali script based on the acoustic signals they perceived. This process is called dialectal writing ~\cite{nigmatulina-etal-2020-asr}, which entails creating phonemic transcriptions that closely align grapheme labels with the acoustic signals, despite their inherent inconsistency. This approach, in our view, mimics what speakers of the varieties would do should they attempt to write them. It took the annotators about four hours to annotate 200 sentences each.
    

    \item \textbf{Griko:} We use a sample of existing Griko (\textit{Italiot Greek}) data~\cite{anastasopoulos-etal-2018-part}. A speaker of both Griko and modern standard Greek created the ``translations'' into modern standard Greek, ending with 163 sentences.

    \item \textbf{Central Kurdish Varieties:} Kurdish is known as a dialect continuum and is mainly classified into Northern, Central, and Southern dialects and is closely related to Zaza-Gorani languages, Laki and Lori ~\cite{ahmadi2023approaches}. In this project, we focus on the varieties of Central Kurdish, also known as Sorani, which are mainly spoken in Kurdistan of Iran, and Iraq. The following local names are generally and broadly used to refer to the dialects of Central Kurdish spoken in regions of the cities specified in parentheses: Babanî (Sulaymaniyah, Iraq) ~\cite{mccarus1956descriptive}, Ardalanî (Sanandaj, Iran), Cafî (Javanrud, Iran), Mukriyanî or Mukrî (Mahabad, Iran) ~\cite{de2018ergin} and Hewlêrî (Erbil, Iraq). Among these, the variant of Sulaymaniyah is the most studied one, which is also widely used as a standard variant of Central Kurdish in the press and media ~\cite{thackston2006sorani}. 
    
    According to various linguistic analyses of fieldwork data,~\citet{matras2019revisiting} classifies Central Kurdish varieties into Northern and Southern Sorani, with their epicenters being based on the dialects of Erbil (\textit{Hewlêr} in Kurdish) and Sulaymaniyah (\textit{Silêmanî} in Kurdish). Based on this classification, Babanî, Ardalanî, and Cafî or Jafi belong to Southern Sorani, while Mukriyanî and Hewlêrî belong to Northern Sorani. Similarly, we believe that the selected varieties can further elucidate the distinctiveness of the varieties and the classification quantitatively.
    
    Given that there are no corpora documenting varieties of Central Kurdish, we resort to movies where speakers of these varieties play a role. To that end, we transcribe movies in Babanî, Ardalanî, and Mukriyanî. Since none of these movies are available in other varieties, we perform a dialect translation by native speakers of Ardalanî, Mukriyanî, and Hewlêrî by randomly selecting and translating 300 sentences in Babanî transcriptions. To mitigate the impact of orthography on the dialect, we normalize and standardize the sentences based on the common orthography of Kurdish using KLPT ~\cite{ahmadi2020klpt}. This way, we create a parallel corpus containing sentences in four dialects of Central Kurdish along with their translations in English. It is worth noting that the collected sentences contain vocabulary of general parlance and capture interesting morphological variations across dialects.
    
    
    \item \textbf{Occitan Varieties:} We focus on two examples of the Occitan continuum, namely Central Occitan and Aranese. We use Central Occitan data from the dissertation of~\cite{dansereau1985studies} who studied the syntax of central Occitan, providing additional translations of all examples to "standard" French (379 sentences). For Aranese (the standardized form of the Pyrenean Gascon variety of Occitan), we scraped a total of 476 sentences from a local news website\footnote{\url{https://web.gencat.cat/en/actualitat/darreres-noticies/index.html}} in Aranese and English. Note that the data in the two varieties are not parallel; thus, we do not have comparable results between these two varieties. We benchmark them for future work.
\end{itemize}

\section{Evaluation}
To assess the quality of any MT system on dialectal variations, it is crucial to compare its outputs with a reference standard. One approach is to have a gold, human-created translation representing the desired translation in a standard setting. Among the twelve languages considered, we only have gold translations for Basque, Bengali, Farsi, Central Kurdish, Malay-Indonesian, Swahili, Tigrinya, and Aranese. For the rest, we will need to be able to evaluate MT robustness without references.

\vspace*{-2mm}
\paragraph{Evaluating Without References}
Our goal is to evaluate the robustness of MT systems concerning dialectal variation. While access to human-created gold translations can certainly reveal a complete picture of the model's performance, thankfully, it is not a hard requirement.

In this work, we adapt the ideas of~\citet{michel-neubig-2018-mtnt} and~\citet{michel-etal-2019-evaluation}
which presented frameworks for evaluating the robustness of MT systems to adversarial or non-native noisy inputs. Concretely, consider the following notation:
\begin{itemize}[noitemsep,nolistsep]
    \item $\mathbf{x}$: the dialectal input sentence.
    \item $\tilde{\mathbf{x}}$: the contrastive sentence  in the "standard" variety. This is deemed to be similar to what MT systems have been trained on and can likely decently translate.
    \item $\mathbf{y}$: the output of the NMT system when $\mathbf{x}$ is provided as input.
    \item $\tilde{\mathbf{y}}$: the output of the NMT system when $\tilde{\mathbf{x}}$ is provided as input.
\end{itemize}

The core of the idea is that we can treat $\tilde{\mathbf{y}}$, the output of the MT system on the "standard" input, as a \textit{pseudo-reference} for the translation. Intuitively, a robust system should produce the same output for inputs with similar meanings regardless of the small dialectal variations.
Hence, we can calculate any MT metric such as BLEU ~\cite{papineni-etal-2002-bleu} or COMET ~\cite{rei-etal-2020-comet} by comparing $\mathbf{y}$ to $\tilde{\mathbf{y}}$.

\vspace*{-2mm}
\paragraph{Important Implementation Notes}
In this work, we focus on two metrics, BLEU and COMET. BLEU compares the $n$-grams of the candidate translation's $n$-grams with the reference translation, counting the number of matches to determine similarity. We calculate BLEU using SacreBLEU ~\cite{post-2018-call}. For space constraints, we do not show the BLEU scores. On the other hand, COMET is a neural framework designed for training multilingual machine translation evaluation models. It leverages information from both the source input and a target-language reference translation to provide more accurate predictions of MT quality, correlating with human judgments. These metrics offer quantitative measures to evaluate and compare the quality of dialectal translations against the reference standards.

Note that both BLEU and COMET are corpus-level scores. For some collections of varieties, though, we have a different number of contrastive sentences ($p$) for a particular dialectal variation compared to the number of standard dialectal sentences ($n$). In such a case, we can still perform individual translations and score each sentence separately. Each contrastive sentence is translated and scored individually using the chosen evaluation metric. Once the scores for all the \emph{p} contrastive sentences are obtained, we calculate an average metric score.

This approach enables us to evaluate the quality of translation on a sentence level. However, a limitation arises from the varying number of \emph{p} for different dialects, resulting in variations in sentence combinations. Consequently, scores cannot be directly compared between dialects. This scenario applies to varieties in four languages: Arabic, Basque, Italian, and Swiss German. To establish comparability, one solution is to create a subset of sentences in all dialects. Unfortunately, the only case where this leads to a decently-sized test set is in Arabic (2000 sentences are shared among all vernaculars). The number of subset sentences among all dialects is presented in Appendix~\ref{tab:subsets}.

We employ an alternative approach for the remaining three languages by selecting a subset of sentences with high dialectal coverage and evaluating the translations exclusively on those dialects. In the case of Basque, we see 34 common sentences among the dialects. Similarly, for Swiss German, we see 87 common sentences. However, for Italian, the data intersection of all varieties is empty. 

We argue that this small number of sentences cannot show the quality appropriately, so we implement an alternative approach for these three languages. First, we exclude dialects that consist of fewer than 100 sentences. This means excluding 50 Italian varieties. Next, for each of the remaining dialects, we randomly select 100 sentences and evaluate the translations based on these samples. We calculate the score for each set of 100 sentences, repeating this process 100 times. Subsequently, we compute the average of the 100 scores obtained from these different runs, representing the final score for that particular dialect.

\section{Results and Analysis}
\paragraph{Preliminaries}
For all language varieties, we benchmark MT systems in the X-to-English direction. The choice of English as a target language is a pragmatic one. Still, a more comprehensive evaluation should consider many other target languages for future work, especially since we do not require gold references to perform our analyses.

We present baseline results in all languages using four different-sized NLLB-200 ~\cite{nllbteam2022language} models using the HuggingFace ~\cite{wolf2020huggingfaces} toolkit. The NLLB-200 can translate between 200 languages. This model has been trained using the teacher-student procedure to work on low-resource languages. To create a large amount of data for NLLB-200 training, the older LASER\footnote{https://github.com/facebookresearch/LASER} (Language-Agnostic SEntence Representation) model was trained on 200 languages. For Italian, we also fine-tune the DeltaLM-large ~\cite{ma2021deltalm} model with Italian-English OPUS ~\cite{tiedemann-2012-parallel} parallel data using the Fairseq ~\cite{ott2019fairseq} toolkit. As we see the superiority of the NLLB models, we do not fine-tune DeltaLM for the rest of the languages.

The COMET evaluation framework relies on XLM-RoBERTa ~\cite{conneau2020unsupervised}, a multilingual language model, to generate embeddings for each token in the input source, machine-translated (mt) sentence, and reference sentence. However, since XLM-RoBERTa was trained on texts of the standard dialect, the quality of the embeddings created for source sentences in different dialectal variants may be compromised. To investigate this, an ablation study was conducted with and without the source sentence as input to the COMET scorer.

\begin{figure*}[!t]
    \centering
    \includegraphics[width=\textwidth]{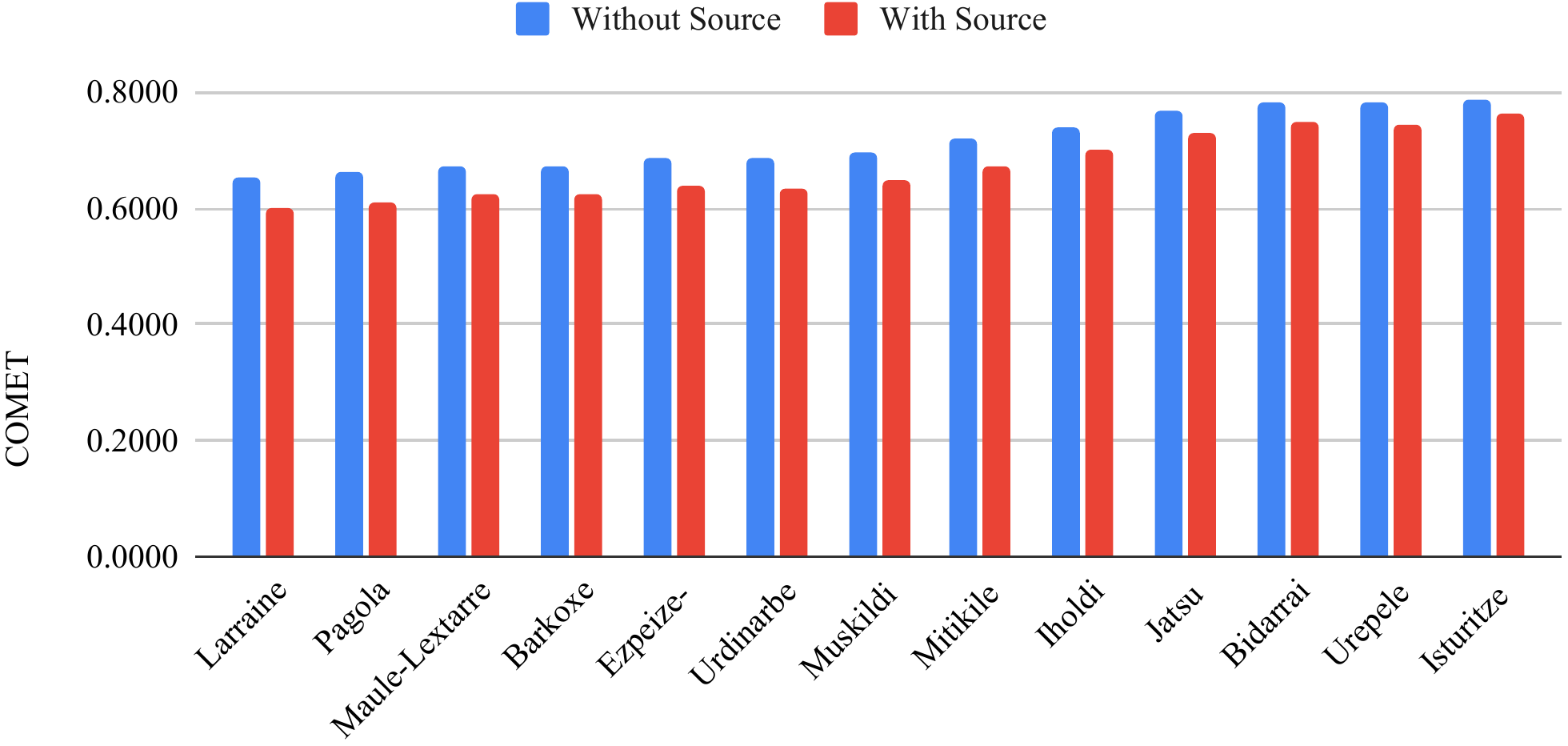}
    \caption{Ablation study of the source sentence usage in dialects of Basque during COMET measurement. COMET scores for Basque varieties when we use the source range from 0.60 to 0.76, but when we don't use the source, they range from 0.65 to 0.79}
    \label{fig:COMET}
\end{figure*}

Figure~\ref{fig:COMET} presents the results of this ablation study for 13 Basque dialects. The dialectal sentences were translated to English using the NLLB-200-dis-600M model. The blue bars represent COMET scores when the source sentences were replaced with blank sentences, while the orange lines represent COMET scores when the source sentences were included. In all cases, the COMET scores decrease when the source sentences are introduced, supporting the initial hypothesis. The general trends are very similar with and without using the source sentence. Based on these findings, the source sentence will not be used to ensure more reliable evaluations for all subsequent COMET calculations in this paper.

\subsection{Quantitative Analysis}
\begin{figure}[t]
    \centering
    \includegraphics[width=0.3\textwidth]{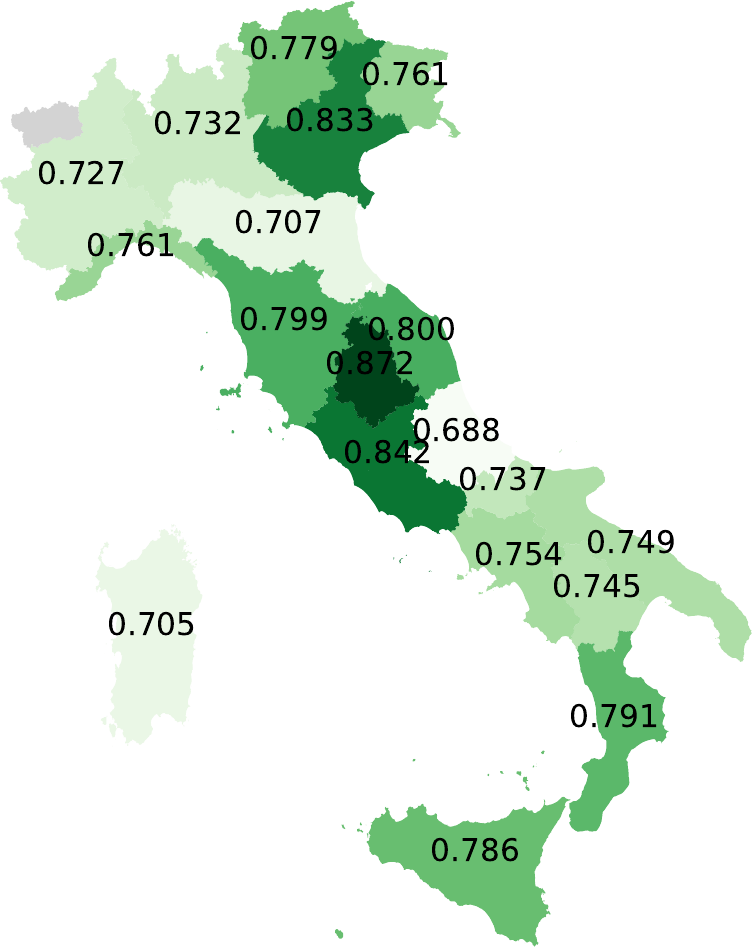}
    \vspace*{-2mm}
    \caption{Map of Italy with COMET scores per region.}
    \label{fig:Italy}
    \vspace*{-6mm}
\end{figure}

\paragraph{Italian Varieties}
The dataset used in this study comprises a total of 439 Italian dialects, which are associated with 290 communes. The COMET scores for four different NLLB-200 models, along with the number of contrastive sentences available for each commune compared to the standard variation, are presented in Table~\ref{tab:italian-communes} in Appendix~\ref{sec:appendix-italaian}. As mentioned earlier, these results are not directly comparable but can be considered a rough estimation of the expected quality. We present the comparable results among all the dialects in Table~\ref{tab:italian-communes-comp} in Appendix \ref{sec:appendix-italaian}.

These 290 communes are further categorized into 78 provinces. Additionally, these 78 provinces are distributed among 19 regions. The comparable COMET scores for these 19 regions can be found in Table~\ref{tab:italian-regions-comp}. We also provide the non-directly-comparable results using all the sentences in Table~\ref{tab:italian-regions-comp} in Appendix~\ref{sec:appendix-italaian}.

Examining the top five COMET scores of the NLLB-Dis-1.3B model, indicated in bold in the Table, it is evident that these dialects strongly resemble the standard variation. This is particularly true for the Tuscany variety, as standard Italian is based on this region. Similarly, the proximity of the other three regions (Umbria, Lazio, Marche) to Tuscany suggests that the similarity of these varieties to the now-standard one is reflected in the MT quality.

Based on the obtained scores, it is possible to visualize them on the map of Italy using geojson information, such as the one available here.\footnote{\url{https://github.com/openpolis/geojson-italy}} Figure~\ref{fig:Italy} illustrates the COMET scores of various regions represented on the map of Italy. A darker shade of green indicates a higher COMET score. The visualization shows that regions near Tuscany are darker green, indicating higher scores. However, the scores gradually decrease as we move further away from those regions.

\begin{figure}[t]
\centering
    \includegraphics[width=0.4\textwidth,height=40mm]{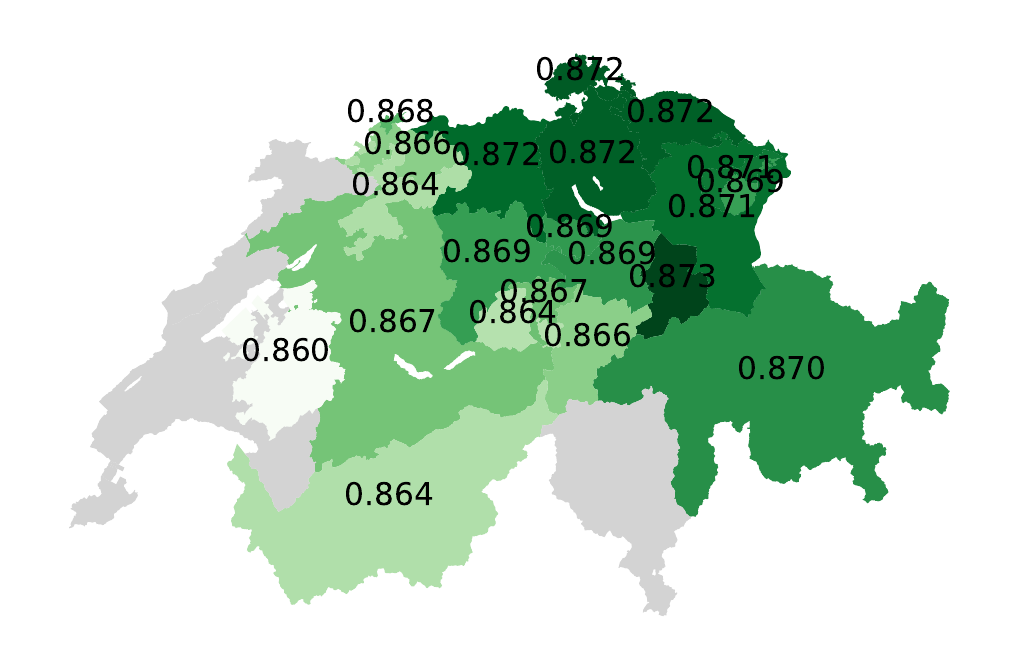}
    \vspace*{-2mm}
    \caption{Map of Switzerland with COMET scores for different regions.}
    \label{fig:Switzerland}
    \vspace*{-6mm}
\end{figure}

\vspace*{-2mm}
\paragraph{Swiss German Varieties}
Similar to the approach taken with Italy, the regional MT quality scores can be geographically visualized on a map. We point the reader to Figure~\ref{fig:Switzerland}, which showcases the map of Switzerland. The map reveals a consistent pattern where the northern regions, being closer to Germany (and consequently speaking varieties closer to High German), obtain higher COMET scores. In contrast, the scores gradually decrease as one moves further south.
Tables~\ref{tab:swiss-dialects} and~\ref{tab:swiss-dialects-comp} present the benchmark scores for Swiss German dialects in non-comparable and comparable formats, respectively. These Tables provide additional valuable information on the dialects and their respective regions. Last, Table~\ref{tab:swiss-regions} and Table~\ref{tab:swiss-regions-comp} in the same appendix display the benchmark scores for different regions of Switzerland in non-comparable and comparable formats, respectively.

\begin{figure*}[t]
    \centering
    \includegraphics[width=.9\textwidth]{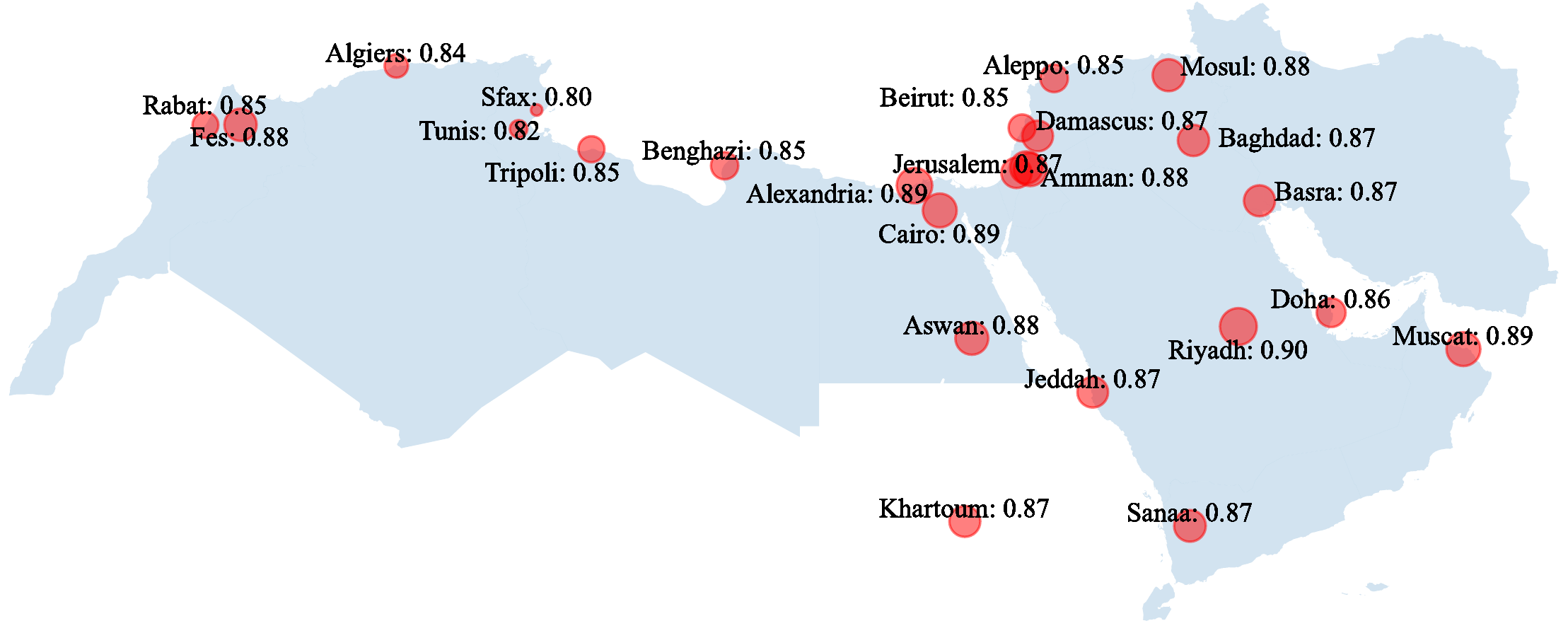}\\
    \vspace*{-2mm}
    \caption{MT quality for Arabic vernaculars. Comet scores range from 0.8 (Sfax, Tunisia) to 0.9 (Riyadh, SA).}
    \label{fig:arabic}
    \vspace*{-2mm}
\end{figure*}
\paragraph{Bengali Varieties}
Table~\ref{tab:other-languages} presents the COMET scores of Bengali across the five varieties. These scores are comparable as they were evaluated using the same 200 sentences. These dialects are spoken in various regions of Bangladesh, and we visualize their distribution on a map in Figure~\ref{fig:Bangladesh}. Interestingly, a similar pattern emerges in this case as well. Jessore, one of the dialects from which standard Bengali originated, exhibits relatively higher COMET scores. Conversely, as we move away from Jessore, the COMET scores gradually decrease, reflecting a relative decline in quality.

\paragraph{Arabic Vernaculars}

In this experiment, we compare a variant to the MSA. Figure~\ref{fig:arabic}, as well as 
Tables~\ref{tab:arabic-dialects} and~\ref{tab:arabic-dialects-comp} showcase the benchmark scores for Arabic vernaculars as spoken in different cities. Focusing on the NLLB-3.3B model, we find that the worst-scoring city is Sfax, Tunisia, and the best-scoring city is Riyadh, Saudi Arabia. The difference is 0.1 COMET point, and all the scores are above 0.8. We can thus infer that the baseline systems represent most Arabic vernaculars fairly well. That said, it is worth noting that the top four scoring cities (Riyadh, Alexandria, Muscat, and Cairo) are close to the Middle East. On the other hand, the bottom no four scoring cities (Sfax, Tunis, Algiers, and Rabat) are all in the West Arab world (in North Africa).

\vspace*{-2mm}
\paragraph{Central Kurdish Varieties}
Table~\ref{tab:other-languages} displays the COMET scores for the different varieties of Central Kurdish, focusing on the dialects spoken in Iran and Iraq. These scores are comparable as they were evaluated using a consistent set of 300 sentences. The geographic distribution of these dialects is worth noting, with Sulaymaniyah located centrally within the region where Central Kurdish is spoken. An intriguing observation is that Sulaymaniyah, situated in the middle of the region, exhibits a higher COMET score. This suggests that the standard variation of Central Kurdish may have emerged from Sulaymaniyah or nearby locations. On the Iraq side, Mahabad stands out with the highest COMET score, indicating its similarity to Sulaymaniyah. The COMET scores gradually drop as we move from these two areas towards the north or south.

Due to space constraints, we provide further quantitative analysis for the other languages in Appendix~\ref{appendix_more_quantitative} with results presented in Table~\ref{tab:other-languages}.

\begin{table*}[t!]
\centering
\resizebox{\textwidth}{!}{
\begin{tabular}{rlrrrrr}\toprule
\multirow{2}{*}{\textbf{Standard}} &\multirow{2}{*}{\textbf{Variety}} &\multirow{2}{*}{\textbf{\# Sentences}} &\multicolumn{4}{c}{\textbf{COMET}} \\\cmidrule{4-7}
\textbf{Language} & & &\textbf{NLLB-Dis-600M} &\textbf{NLLB-Dis-1.3B} &\textbf{NLLB-1.3B} &\textbf{NLLB-3.3B} \\\midrule
\multirow{2}{*}{\textbf{Tigrinya}} &\textbf{Ethiopian} &3071 &0.8017 &0.8232 &0.8173 &0.8245 \\
&\textbf{Eritrean} &3071 &0.7782 &0.7998 &0.7972 &0.8039 \\\cmidrule{2-7}
\multirow{2}{*}{\textbf{Farsi}} &\textbf{Farsi} &3071 &0.8458 &0.8545 &0.8532 &0.8564 \\
&\textbf{Dari} &3071 &0.8387 &0.8494 &0.8480 &0.8539 \\\cmidrule{2-7}
\multirow{2}{*}{\textbf{Malay-Indonesian}} &\textbf{Indonesian} &3071 &0.8608 &0.8666 &0.7407 &0.7330 \\
&\textbf{Malay} &3071 &0.8542 &0.8625 &0.8077 &0.7965 \\\cmidrule{2-7}
\multirow{2}{*}{\textbf{Swahili}} &\textbf{Coastal} &1991 &0.8508 &0.8622 &0.8611 &0.8657 \\
&\textbf{Congolese} &1991 &0.8094 &0.8253 &0.8206 &0.8229 \\\cmidrule{2-7}
\multirow{2}{*}{\textbf{Occitan}} &\textbf{Aranese} &476 &0.7537 &0.7743 &0.7752 &0.7841 \\\cmidrule{2-7}
&\textbf{Central} &379 &0.7050 &0.7400 &0.7425 &0.5439 \\\cmidrule{2-7}
\multirow{4}{*}{\textbf{Central Kurdish}} &\textbf{Sulaymaniyah} &300 &0.7295 &0.7427 &0.7419 &0.7436 \\
&\textbf{Erbil} &300 &0.6975 &0.7133 &0.7099 &0.7167 \\
&\textbf{Sanandaj} &300 &0.6763 &0.6941 &0.6916 &0.6969 \\
&\textbf{Mahabad} &300 &0.7201 &0.7348 &0.7237 &0.7351 \\\cmidrule{2-7}
\multirow{5}{*}{\textbf{Bengali}} &\textbf{Barisal} &200 &0.7038 &0.7089 &0.7176 &0.7266 \\
&\textbf{Dhakaiya} &200 &0.7876 &0.8006 &0.7969 &0.8012 \\
&\textbf{Jessore} &200 &0.8226 &0.8395 &0.8332 &0.8365 \\
&\textbf{Khulna} &200 &0.8121 &0.8193 &0.8241 &0.8295 \\
&\textbf{Kushtia} &200 &0.7922 &0.7992 &0.8144 &0.8132 \\\cmidrule{2-7}
\textbf{Greek} &\textbf{Griko} &163 &0.4877 &0.4969 &0.4964 &0.5065 \\
\bottomrule
\end{tabular}
}
\vspace*{-2mm}
\caption{COMET scores of different languages' dialects for various model scales. There often exist significant differences between the varieties. Bigger models are better than smaller ones, but dialectal inequalities persist.}
\label{tab:other-languages}
\vspace*{-4mm}
\end{table*}
\vspace*{-2mm}

\subsection{Qualitative Analysis}
One of the major factors that affect the performance of NMT systems when dealing with dialects is the various lexical and morpho-syntactic variations among dialects and varieties. The standardization process of a language culminates in establishing linguistic homogeneity within its vocabulary, often to the detriment of regional dialects or linguistic varieties.
We posit that the inadequate lexical representation of nonstandard dialects has a detrimental impact on the performance of NMT systems, including pre-trained ones.

Moreover, some selected languages, like Kurdish, spoken in different countries, deal with code-switching phenomena more prevalent than others due to socio-linguistic factors. This is particularly the case of loanwords and terminologies. For instance, words that pertain to automobile mechanics in the Kurdish spoken in Iran are mostly borrowed from Russian while the Kurdish spoken in Iraq relies more on the Arabic and English words in this domain. In the same vein, standard orthographies, if they exist for a language, implicitly create a bias in transcription and inaccuracy in translating vernaculars. Since this is not peculiar to the selected languages, we believe it affects NMT systems. 

\begin{table}[!h]
    \includegraphics[width=1\columnwidth]{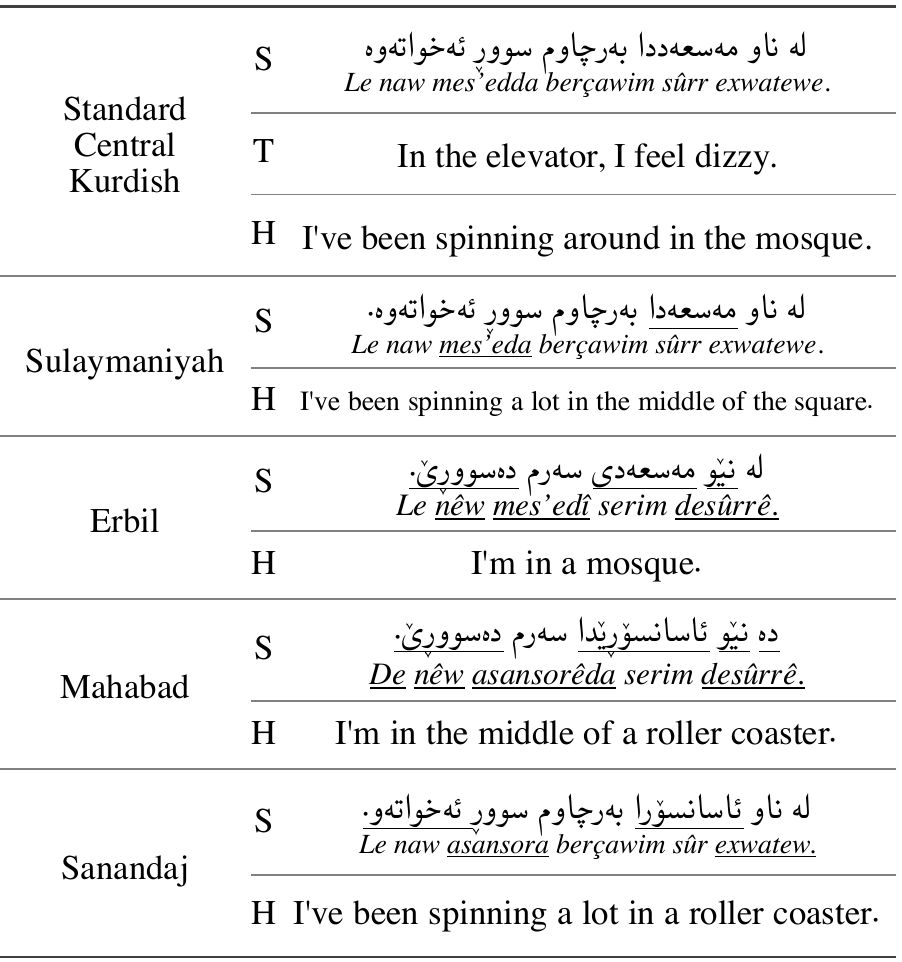}
    \vspace{-7mm}
    \caption{A sentence (S) in Central Kurdish along with transliterations and translations (T) for the dialects in \dataset. Underlined words specify morphosyntactic or lexical variations. H is the MT hypothesis.}
    \label{tab_qualitative_example}
    \vspace{-6mm}
\end{table}

Table~\ref{tab_qualitative_example} shows example translations from our Central Kurdish data in comparison to the dialects in \dataset. On the source side, the underlined morphosyntactic and lexical variations include the postposition `\textit{da}' marking locative case, the word for `elevator', and the compound verb. 

\section{Conclusion}

This study compiles a benchmark of contrastive examples between standard and dialectal variants of twelve languages to facilitate the evaluation of MT systems' robustness along this variation. Our findings demonstrate that MT systems excel at handling standard variants, but as the dialectal varieties start differing from the standard, the quality of the translations declines. This work emphasizes the need for further research and development in dialectal MT. Excluding a significant portion of the population from the benefits of language translation cannot be considered a satisfactory solution, underscoring the importance of addressing dialectal variations within MT systems.

\paragraph{Future Work} This study highlights the unequal support for different language dialects in MT systems. Some dialects exhibit impressive COMET scores due to their close relationship with the standard variant. However, this work primarily focuses on creating a dataset to assess the performance of various dialects rather than conducting experiments to enhance the MT system's robustness. This limitation primarily stems from the scarcity of training data. The datasets created for this study are relatively small and mainly serve as test data.

For future research, the MT community needs to prioritize the development of training datasets for dialects. Several strategies can be explored with an adequate dataset, such as dialect-specific adaptation through fine-tuning or adapter approaches.

\section{Limitations}
One of the limitations of our study is the lack of classification which can describe the expected levels of dissimilarity across dialects of a given language. Such a classification can provide the words and labels that are used to denote each dialect. This, however, is not an easy task given the different classifications and various names used for dialects. On the other hand, we believe that other factors that determine the performance of NMT systems should be further studied in regard to dialects. 

\section*{Acknowledgments}
This work was generously supported by the National Science Foundation under awards IIS-2125466 and BCS-2109578, and a Sponsored Research Award from Meta. The authors are also grateful to everyone who contributed to the resources to create the dataset, as well as to the Office of Research Computing at GMU, where all computational experiments were conducted. Sina Ahmadi acknowledges support of the Swiss National Science Foundation (MUTAMUR; no. 213976).

\bibliography{anthology,custom}

\newpage
\appendix

\counterwithin{figure}{section}
\counterwithin{table}{section}

\section{The Languages of \dataset}
\label{app:languages}

\paragraph{Basque Varieties}
Our Basque data is sourced from the Basque Syntactic Database.\footnote{\url{http://ixa2.si.ehu.eus/atlas2/index.php}} 
To gather and analyze the data, researchers initially developed specific questionnaires, each focusing on a distinct linguistic phenomenon characterized by syntactic variation, for a total of  370 different questions. These questionnaires were then provided to informants spanning different age groups, carefully selected from various locations, which comprise 39 variants in the Northern Basque Country in France. 

By posing identical questions to speakers of different Basque dialects, this methodology creates contrastive data facilitating an $n$-way comparison among the dialects. One challenge encountered in this process is that the questions themselves are presented in French. Consequently, we lack sentences in the standard variant. This said, the provided English translations of French sentences serve as gold-standard reference translations.

\paragraph{Italian Varieties and Languages}
Our Italian data are obtained from the Italian Syntactic Atlas\footnote{\url{http://svrims2.dei.unipd.it:8080/asit-maldura/pages/search.jsp}} which functions similarly to the Basque one. However, in the Italian Syntactic Atlas, the questions are presented in standard Italian. This extensive dataset consists of 792 questions that speakers of various Italian dialects have answered. The dataset encompasses a rich collection of 439 dialects from different regions across Italy. Additionally, the dataset provides information about the specific locations where these dialects are spoken. This comprehensive resource enables in-depth analysis and exploration of the dialectal variations found within the Italian language.

It is important to note that many of the vernaculars spoken around Italy are recognized as officially distinct languages (e.g., Neapolitan, Ligurian, and Venetian, to name a few). Some of these also have a distinct online presence (e.g., with decent Wikipedias), and some MT research is devoted to them ~\cite{nllbteam2022language}. However, this "discretization" of the language continuum observed in the Italian peninsula, where each city/village is said to have its dialect, is far from realistic. Hence we focus on the fine-grained evaluation that our data from over 439 locales allows.

\paragraph{Swiss German Varieties}
Our Swiss German data was obtained by scraping the Syntactic Atlas of German Switzerland (SADS).\footnote{\url{https://dialektsyntax.linguistik.uzh.ch}} 
The SADS website hosts a total of 118 questionnaires, each accompanied by answers provided in 368 different locales. This dataset allows for an $n$-way comparison between the dialects and the standard (Swiss) German variant, providing valuable contrastive information. However, the data available on the website primarily focuses on highlighting the changes present in the sentences, necessitating manual annotation to identify instances where alterations occur in standard German sentences. Through this manual annotation process, we captured the specific linguistic variations exhibited by the Swiss German dialects.

\paragraph{Central Occitan and Aranese}
Occitan is a Romance language spoken in southern France, Monaco, Italy, and Catalonia, also known as Provençal or Languedocian (\textit{lange d'oc}), and acknowledged as a language continuum with multiple variations. In this work, we use data from the dissertation of ~\cite{dansereau1985studies} who studied the syntax of central Occitan, providing additional translations of all examples to "standard" French. In total, we have 379 in the Occitan portion of \dataset. Note, of course, that French and Occitan are widely accepted as different languages; nevertheless, most Occitan speakers live in France, and therefore most systems will direct these speakers' input to a French model.

Aranese is a standardized form of the Pyrenean Gascon variety of the Occitan language. It is primarily spoken in the Val d'Aran, located in northwestern Catalonia near the border between Spain and France. Aranese holds official status alongside Catalan and Spanish as one of the three recognized languages in this region. In our research, we scraped a total of 476 sentences from the gencat website,\footnote{\url{https://web.gencat.cat/en/actualitat/darreres-noticies/index.html}} in Aranese and English. 

\paragraph{Griko} Griko is a Greek dialect spoken in southern Italy, in the Grec\`ia Salentina area southeast of Lecce. It is also known as \textit{Italiot Greek} when combined with the Greko variety of Calabria. For \dataset, we use a sample of Griko data from ~\cite{anastasopoulos-etal-2018-part}, for which we also create ``translations'' into modern standard Greek, ending up with a total of 163 sentences.

\paragraph{Arabic Vernaculars}
Arabic, as a macro-language, encompasses a range of dialects within its language continuum. Modern Standard Arabic (MSA) is a standardized form of the language used across various regions, encompassing cultural, media, and educational domains from Morocco to the west to Oman to the east. However, it is important to note that MSA is not the native language of Arabic speakers. In informal and spontaneous settings where spoken MSA is typically expected, such as in TV talk shows, speakers often code-switch between their respective vernaculars and MSA.

To examine MT performance in Arabic dialects, we use the MADAR corpus ~\cite{bouamor-etal-2018-madar}. This extensive corpus consists of 12000 sentences on varieties from 25 different Arabic-speaking cities. The corpus is created by translating selected sentences from the Basic Traveling Expression Corpus (BTEC)~\cite{takezawa-etal-2007-multilingual} into various dialects and MSA. This unique dataset is highly suitable for conducting contrastive machine translation (MT) research for Arabic dialects, but to our knowledge has not been extensively used for this purpose.

\paragraph{Tigrinya}
Tigrinya is an Ethio-Semitic language predominantly spoken in Eritrea and by the Tigrayan people in the Tigray Region of northern Ethiopia. Within Tigrinya, two major varieties exist the Eritrean dialect and the Ethiopian dialect. To explore and compare these two, we leverage the dataset available from TICO-19~\cite{anastasopoulos-etal-2020-tico}. The TICO-19 dataset is the result of a collective translation initiative during the COVID-19 pandemic, aiming to enhance society's readiness to respond to the ongoing crisis through the utilization of translation technologies effectively. This dataset specifically focuses on the COVID-19 domain, containing translations of the same content in multiple languages. The same 3071 English sentences were professionally translated into both varieties of Togrinya, making it ideal for our purposes.

\paragraph{Farsi and Dari}
We use the same TICO-19 dataset to obtain the data we need for Farsi as spoken in Iran and one of its variants, Dari, as spoken in Afghanistan. 7.6 million people speak Dari. These 2 languages are mutually intelligible in written format but very different when spoken.

\paragraph{Malay and Indonesian}
The TICO-19 dataset also provides data in Malay and one of its standardized variants, Indonesian. Malay serves as the official language in Brunei, Indonesia, Malaysia, and Singapore, and it is also spoken in East Timor, parts of the Philippines, and Thailand. Overall, Malay is spoken by approximately 290 million individuals. Out of this total, the Indonesian variant is spoken by around 260 million people in Indonesia. Though both languages are generally mutually intelligible, the spelling, grammar, pronunciation, vocabulary, and source of loanwords make a noticeable difference between them.

\paragraph{Swahili} We use the Coastal and Congolese Swahili data produced by the TICO-19 dataset, as before. The two varieties are largely intelligible, although the Coastal one (spoken in Tanzania and Kenya) has more influences from English, while the Congolese one incorporates more elements from French.

\section{Quantitative Analysis}
\label{appendix_more_quantitative}

\paragraph{Basque Varieties}
Tables~\ref{tab:basque-dialects} and~\ref{tab:basque-dialects-comp} contain the benchmark scores for Basque dialects.\footnote{Due to space constraints, these results are provided in the Appendix~\ref{app:results_appendix}.} The lowest-scoring dialect is Maule-Lextarre, and the highest-scoring one is Urruna, with a difference of around 0.15 COMET points. This shows that further work is needed for a good MT system for under-represented dialects.

\paragraph{Other Languages}
Table~\ref{tab:other-languages} displays the results for all the other languages\footnote{In Appendix~\ref{app:results_appendix}, we present the benchmark results for all languages.} encompassing only 1-3 dialects. As for Griko, Central Occitan, and Aranese, we have no other dialects to compare their results. Nevertheless, we benchmark them for future work. We base our discussion below on the best-performing NLLB-3.3B model.

For Tigrinya, the Ethiopian dialect has a higher COMET score (0.82) than the Eritrean dialect (0.8). This is consistent for all pre-trained models. Even though Tigrinya is the largest language of Eritrea (unlike Ethiopia), the model seems better suited to the Ethiopian dialect -- we suspect this is because most online resources are in this variety.

Regarding Farsi and Dari, the pre-trained models perform almost equally well despite a small difference between these two dialects (around 0.01 COMET points on average).
For Malay-Indonesian, the results are more mixed. The distilled models obtain better COMET scores for Indonesian than Malay in general. This may be expected because the NLLB models support Indonesian but not Malay. However, we observe an opposite trend for the two non-distilled models, where the Malay language gets a higher COMET score.

For Swahili, the result is consistent for all the pre-trained models: Coastal variety is better handled than Congolese. The Coastal variety is highly resourced and included in the models' training, unlike the Congolese one, which is primarily spoken.

\begin{figure*}[!t]
    \centering
    \includegraphics[scale=0.75]{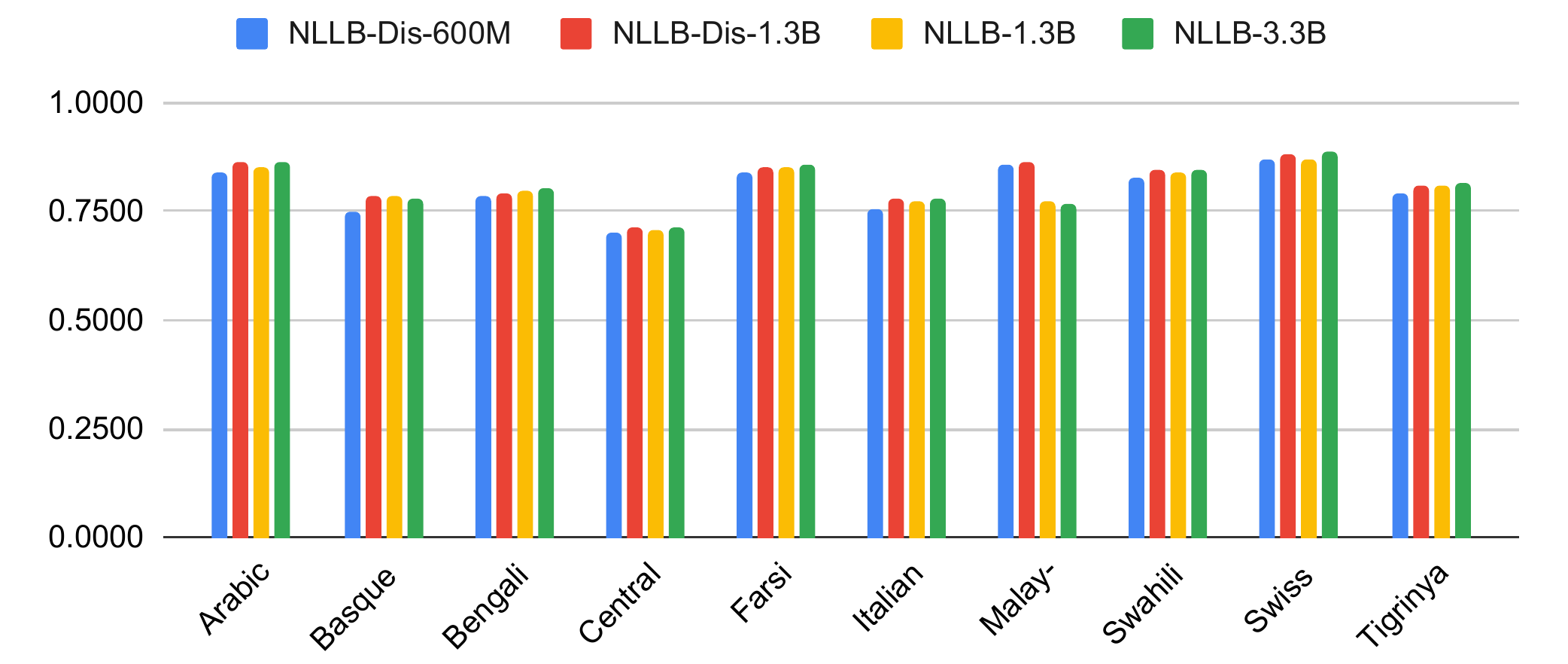}
    \caption{Average COMET score of all the dialects of languages with more than one variety.}
    \label{fig:avg_COMET}
\end{figure*}

Comparing average results across languages (Figure~\ref{fig:avg_COMET} depicts the average COMET scores), we find that the baseline system performs well for the various dialects of Swiss German, Farsi, and Arabic but not as well for other languages, especially low-resourced ones. Comparing the models based on size, we find that larger ones consistently outperformed the smaller ones.

\section{Complete Results}
\label{app:results_appendix}

\begin{figure*}[t]
    \centering
    \includegraphics[width=0.3\textwidth]{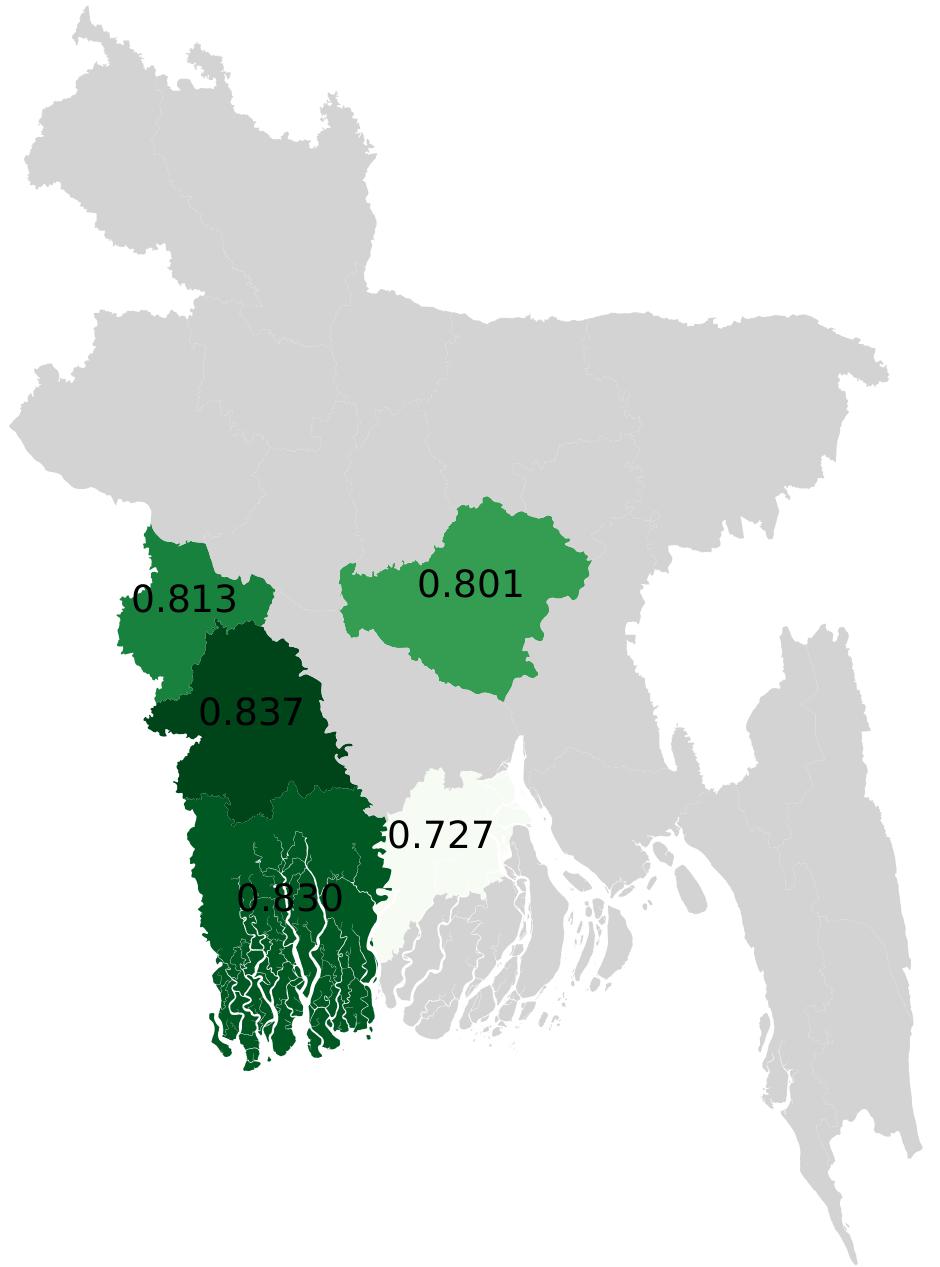}
    \caption{Map of Bangladesh with COMET scores for different regions.}
    \label{fig:Bangladesh}
\end{figure*}

\begin{table*}[!t]
    \centering
\begin{tabular}{l|c|c}
    \toprule
    \multirow{2}{*}{\textbf{Language}} & \multirow{2}{*}{\textbf{\# Sentences}} & \multirow{2}{*}{\textbf{\# Sentences}} \\
    & \textbf{(common)} & \textbf{(coverage)} \\ \midrule
    Arabic & 2000 & \\
    Basque & 0 & 34\\
    Italian & 0 & \\
    Swiss German & 0 & 87\\
    \bottomrule
    \end{tabular}
    \caption{The subset of common sentences and those with the highest coverage in all dialects of the indicated languages. Except for Arabic, there is no common sentence for the other languages. }
    \label{tab:subsets}
\end{table*}

\label{sec:appendix-arabic}
\begin{table*}[t!]
\centering
\resizebox{0.83\textwidth}{!}{%
\begin{tabular}{l|r|rrrrr|}\toprule
\multirow{2}{*}{\textbf{Arabic}} &\multirow{2}{*}{\textbf{\# of Sentences}} &\multicolumn{4}{c}{\textbf{COMET}} \\\cmidrule{3-6}
& &\textbf{NLLB-Dis-600M} &\textbf{NLLB-Dis-1.3B} &\textbf{NLLB-1.3B} &\textbf{NLLB-3.3B} \\\midrule
\textbf{Alexandria} &2000 &0.8655 &0.8895 &0.8811 &0.8947 \\
\textbf{Baghdad} &2000 &0.8445 &0.8649 &0.8595 &0.8711 \\
\textbf{Doha} &12000 &0.8380 &0.8572 &0.8509 &0.8588 \\
\textbf{Benghazi} &2000 &0.8336 &0.8496 &0.8452 &0.8520 \\
\textbf{Khartoum} &2000 &0.8488 &0.8656 &0.8626 &0.8695 \\
\textbf{Sfax} &2000 &0.7815 &0.8015 &0.7990 &0.8010 \\
\textbf{Muscat} &2000 &0.8639 &0.8839 &0.8790 &0.8855 \\
\textbf{Mosul} &2000 &0.8430 &0.8649 &0.8619 &0.8753 \\
\textbf{Riyadh} &2000 &0.8859 &0.9011 &0.8966 &0.9028 \\
\textbf{Sanaa} &2000 &0.8452 &0.8704 &0.8633 &0.8733 \\
\textbf{Aswan} &2000 &0.8496 &0.8736 &0.8680 &0.8800 \\
\textbf{Algiers} &2000 &0.8162 &0.8330 &0.8276 &0.8357 \\
\textbf{Tripoli} &2000 &0.8271 &0.8406 &0.8380 &0.8465 \\
\textbf{Jeddah} &2000 &0.8420 &0.8653 &0.8615 &0.8683 \\
\textbf{Rabat} &12000 &0.8181 &0.8366 &0.8318 &0.8418 \\
\textbf{Cairo} &12000 &0.8578 &0.8805 &0.8735 &0.8839 \\
\textbf{Jerusalem} &2000 &0.8450 &0.8632 &0.8559 &0.8666 \\
\textbf{Beirut} &12000 &0.8315 &0.8553 &0.8391 &0.8512 \\
\textbf{Basra} &2000 &0.8436 &0.8640 &0.8575 &0.8700 \\
\textbf{Tunis} &12000 &0.7931 &0.8134 &0.8061 &0.8152 \\
\textbf{Damascus} &2000 &0.8457 &0.8660 &0.8545 &0.8686 \\
\textbf{Salt} &2000 &0.8569 &0.8767 &0.8650 &0.8772 \\
\textbf{Fes} &2000 &0.8594 &0.8750 &0.8695 &0.8769 \\
\textbf{Aleppo} &2000 &0.8311 &0.8518 &0.8389 &0.8537 \\
\textbf{Amman} &2000 &0.8618 &0.8767 &0.8683 &0.8811 \\
\bottomrule
\end{tabular}
}
\caption{COMET score of different Arabic dialects on all sentences.}
\label{tab:arabic-dialects}
\end{table*}

\begin{table*}[!t]
\centering
\resizebox{0.83\textwidth}{!}{%
\begin{tabular}{l|r|rrrrr|}\toprule
\multirow{2}{*}{\textbf{Arabic}} &\multirow{2}{*}{\textbf{\# of Sentences}} &\multicolumn{4}{c}{\textbf{COMET}} \\\cmidrule{3-6}
& &\textbf{NLLB-Dis-600M} &\textbf{NLLB-Dis-1.3B} &\textbf{NLLB-1.3B} &\textbf{NLLB-3.3B} \\\midrule
\textbf{Sfax} &2000 &0.7815 &0.8015 &0.7990 &0.8010 \\
\textbf{Tunis} &2000 &0.7942 &0.8124 &0.8062 &0.8159 \\
\textbf{Algiers} &2000 &0.8162 &0.8330 &0.8276 &0.8357 \\
\textbf{Rabat} &2000 &0.8205 &0.8400 &0.8358 &0.8457 \\
\textbf{Tripoli} &2000 &0.8271 &0.8406 &0.8380 &0.8465 \\
\textbf{Beirut} &2000 &0.8285 &0.8518 &0.8363 &0.8503 \\
\textbf{Benghazi} &2000 &0.8336 &0.8496 &0.8452 &0.8520 \\
\textbf{Aleppo} &2000 &0.8311 &0.8518 &0.8389 &0.8537 \\
\textbf{Doha} &2000 &0.8389 &0.8591 &0.8520 &0.8595 \\
\textbf{Jerusalem} &2000 &0.8450 &0.8632 &0.8559 &0.8666 \\
\textbf{Jeddah} &2000 &0.8420 &0.8653 &0.8615 &0.8683 \\
\textbf{Damascus} &2000 &0.8457 &0.8660 &0.8545 &0.8686 \\
\textbf{Khartoum} &2000 &0.8488 &0.8656 &0.8626 &0.8695 \\
\textbf{Basra} &2000 &0.8436 &0.8640 &0.8575 &0.8700 \\
\textbf{Baghdad} &2000 &0.8445 &0.8649 &0.8595 &0.8711 \\
\textbf{Sanaa} &2000 &0.8452 &0.8704 &0.8633 &0.8733 \\
\textbf{Mosul} &2000 &0.8430 &0.8649 &0.8619 &0.8753 \\
\textbf{Fes} &2000 &0.8594 &0.8750 &0.8695 &0.8769 \\
\textbf{Salt} &2000 &0.8569 &0.8767 &0.8650 &0.8772 \\
\textbf{Aswan} &2000 &0.8496 &0.8736 &0.8680 &0.8800 \\
\textbf{Amman} &2000 &0.8618 &0.8767 &0.8683 &0.8811 \\
\textbf{Cairo} &2000 &0.8583 &0.8790 &0.8724 &0.8853 \\
\textbf{Muscat} &2000 &0.8639 &0.8839 &0.8790 &0.8855 \\
\textbf{Alexandria} &2000 &0.8655 &0.8895 &0.8811 &0.8947 \\
\textbf{Riyadh} &2000 &0.8859 &0.9011 &0.8966 &0.9028 \\
\bottomrule
\end{tabular}
}
\caption{Comparable COMET score of different Arabic dialects on a subset of 2000 sentences.}
\label{tab:arabic-dialects-comp}
\end{table*}

\begin{table*}[t!]
\centering
\resizebox{0.83\textwidth}{!}{%
\begin{tabular}{l|r|rrrrr|}\toprule
\multirow{2}{*}{\textbf{Arabic}} &\multirow{2}{*}{\textbf{\# of Sentences}} &\multicolumn{4}{c}{\textbf{BLEU}} \\\cmidrule{3-6}
& &\textbf{NLLB-Dis-600M} &\textbf{NLLB-Dis-1.3B} &\textbf{NLLB-1.3B} &\textbf{NLLB-3.3B} \\\midrule
\textbf{Sfax} &2000 &21.48 &24.11 &23.80 &24.53 \\
\textbf{Tunis} &12000 &23.75 &26.87 &25.76 &27.28 \\
\textbf{Algiers} &2000 &25.20 &28.11 &27.84 &28.91 \\
\textbf{Rabat} &12000 &28.21 &32.13 &31.45 &33.03 \\
\textbf{Tripoli} &2000 &28.48 &32.38 &32.32 &33.70 \\
\textbf{Beirut} &12000 &29.65 &35.53 &32.10 &34.44 \\
\textbf{Benghazi} &2000 &30.72 &35.11 &34.06 &35.68 \\
\textbf{Aleppo} &2000 &30.17 &34.92 &32.86 &36.36 \\
\textbf{Doha} &12000 &31.04 &35.76 &34.75 &36.37 \\
\textbf{Jerusalem} &2000 &31.40 &36.22 &34.55 &37.87 \\
\textbf{Jeddah} &2000 &31.29 &36.33 &35.32 &37.70 \\
\textbf{Damascus} &2000 &31.29 &36.85 &34.58 &38.49 \\
\textbf{Khartoum} &2000 &35.84 &40.19 &39.99 &42.18 \\
\textbf{Basra} &2000 &32.34 &36.84 &35.83 &39.02 \\
\textbf{Baghdad} &2000 &32.71 &37.26 &37.03 &40.04 \\
\textbf{Sanaa} &2000 &32.25 &38.68 &37.18 &39.67 \\
\textbf{Mosul} &2000 &33.16 &39.32 &38.07 &41.44 \\
\textbf{Fes} &2000 &34.77 &39.04 &38.44 &40.90 \\
\textbf{Salt} &2000 &35.12 &41.15 &38.32 &41.56 \\
\textbf{Aswan} &2000 &31.60 &38.29 &36.61 &39.61 \\
\textbf{Amman} &2000 &33.29 &38.55 &36.35 &40.30 \\
\textbf{Cairo} &12000 &33.60 &40.22 &38.41 &41.17 \\
\textbf{Muscat} &2000 &37.01 &43.10 &42.29 &44.13 \\
\textbf{Alexandria} &2000 &36.19 &43.19 &40.51 &44.98 \\
\textbf{Riyadh} &2000 &40.48 &46.55 &45.03 &47.60 \\
\bottomrule
\end{tabular}
}
\caption{BLEU score of different Arabic dialects on all sentences.}
\label{tab:arabic-dialects_BLEU}
\end{table*}

\begin{table*}[!t]
\centering
\resizebox{0.83\textwidth}{!}{%
\begin{tabular}{l|r|rrrrr|}\toprule
\multirow{2}{*}{\textbf{Arabic}} &\multirow{2}{*}{\textbf{\# of Sentences}} &\multicolumn{4}{c}{\textbf{BLEU}} \\\cmidrule{3-6}
& &\textbf{NLLB-Dis-600M} &\textbf{NLLB-Dis-1.3B} &\textbf{NLLB-1.3B} &\textbf{NLLB-3.3B} \\\midrule
\textbf{Sfax} &2000 &21.48 &24.11 &23.80 &24.53 \\
\textbf{Tunis} &2000 &24.31 &27.73 &25.97 &28.13 \\
\textbf{Algiers} &2000 &25.20 &28.11 &27.84 &28.91 \\
\textbf{Rabat} &2000 &29.32 &32.93 &32.47 &33.99 \\
\textbf{Tripoli} &2000 &28.48 &32.38 &32.32 &33.70 \\
\textbf{Beirut} &2000 &29.34 &34.91 &31.78 &34.83 \\
\textbf{Benghazi} &2000 &30.72 &35.11 &34.06 &35.68 \\
\textbf{Aleppo} &2000 &30.17 &34.92 &32.86 &36.36 \\
\textbf{Doha} &2000 &32.05 &36.71 &35.30 &37.64 \\
\textbf{Jerusalem} &2000 &31.40 &36.22 &34.55 &37.87 \\
\textbf{Jeddah} &2000 &31.29 &36.33 &35.32 &37.70 \\
\textbf{Damascus} &2000 &31.29 &36.85 &34.58 &38.49 \\
\textbf{Khartoum} &2000 &35.84 &40.19 &39.99 &42.18 \\
\textbf{Basra} &2000 &32.34 &36.84 &35.83 &39.02 \\
\textbf{Baghdad} &2000 &32.71 &37.26 &37.03 &40.04 \\
\textbf{Sanaa} &2000 &32.25 &38.68 &37.18 &39.67 \\
\textbf{Mosul} &2000 &33.16 &39.32 &38.07 &41.44 \\
\textbf{Fes} &2000 &34.77 &39.04 &38.44 &40.90 \\
\textbf{Salt} &2000 &35.12 &41.15 &38.32 &41.56 \\
\textbf{Aswan} &2000 &31.60 &38.29 &36.61 &39.61 \\
\textbf{Amman} &2000 &33.29 &38.55 &36.35 &40.30 \\
\textbf{Cairo} &2000 &34.30 &40.96 &39.37 &41.86 \\
\textbf{Muscat} &2000 &37.01 &43.10 &42.29 &44.13 \\
\textbf{Alexandria} &2000 &36.19 &43.19 &40.51 &44.98 \\
\textbf{Riyadh} &2000 &40.48 &46.55 &45.03 &47.60 \\
\bottomrule
\end{tabular}
}
\caption{Comparable BLEU score of different Arabic dialects on a subset of 2000 sentences.}
\label{tab:arabic-dialects-comp_BLEU}
\end{table*}
\label{sec:appendix-basque}
\begin{table*}[t!]
\centering
\resizebox{\textwidth}{!}{%
\begin{tabular}{l|r|rrrrr|}\toprule
\multirow{2}{*}{\textbf{Basque}} &\multirow{2}{*}{\textbf{\# of Sentences}} &\multicolumn{4}{c}{\textbf{COMET}} \\\cmidrule{3-6}
& &\textbf{NLLB-Dis-600M} &\textbf{NLLB-Dis-1.3B} &\textbf{NLLB-1.3B} &\textbf{NLLB-3.3B} \\\midrule
\textbf{Ahetze} &197 &0.8045 &0.8058 &0.8073 &0.8050 \\
\textbf{Amenduze-Unaso} &198 &0.8109 &0.8111 &0.8180 &0.8095 \\
\textbf{Arbona} &196 &0.8188 &0.8032 &0.8168 &0.8056 \\
\textbf{Azkaine} &198 &0.8276 &0.8279 &0.8314 &0.8225 \\
\textbf{Baigorri} &198 &0.8009 &0.8088 &0.8070 &0.7961 \\
\textbf{Barkoxe} &198 &0.6728 &0.7014 &0.6904 &0.6878 \\
\bottomrule
\end{tabular}
}
\caption{COMET score of different Basque dialects on all sentences.}
\label{tab:basque-dialects}
\end{table*}

\begin{table*}[t!]
\centering
\resizebox{0.9\textwidth}{!}{%
\begin{tabular}{l|rrrr|}\toprule
\multirow{2}{*}{\textbf{Basque}} &\multicolumn{4}{c}{\textbf{COMET}} \\\cmidrule{2-5}
&\textbf{NLLB-Dis-600M} &\textbf{NLLB-Dis-1.3B} &\textbf{NLLB-1.3B} &\textbf{NLLB-3.3B} \\\midrule
\textbf{Luhuso} &0.7894 &0.8278 &0.8236 &0.8202 \\
\textbf{Jutsi} &0.7863 &0.8144 &0.8218 &0.8173 \\
\textbf{Muskildi} &0.6499 &0.7165 &0.7065 &0.7011 \\
\textbf{Donibane-Lohizune} &0.7713 &0.8102 &0.8032 &0.7982 \\
\textbf{Uharte-Garazi} &0.7636 &0.7877 &0.8008 &0.7977 \\
\textbf{Maule-Lextarre} &0.6254 &0.6949 &0.6723 &0.6816 \\
\bottomrule
\end{tabular}
}
\caption{Comparable COMET score of different Basque dialects}
\label{tab:basque-dialects-comp}
\end{table*}

\begin{table*}[t!]
\centering
\resizebox{\textwidth}{!}{%
\begin{tabular}{l|r|rrrrr|}\toprule
\multirow{2}{*}{\textbf{Basque}} &\multirow{2}{*}{\textbf{\# of Sentences}} &\multicolumn{4}{c}{\textbf{BLEU}} \\\cmidrule{3-6}
& &\textbf{NLLB-Dis-600M} &\textbf{NLLB-Dis-1.3B} &\textbf{NLLB-1.3B} &\textbf{NLLB-3.3B} \\\midrule
\textbf{Luhuso} &198 &21.61 &19.79 &21.06 &19.52 \\
\textbf{Jutsi} &198 &21.30 &19.54 &20.09 &19.85 \\
\textbf{Muskildi} &184 &9.57 &8.04 &9.30 &8.40 \\
\textbf{Donibane-Lohizune} &198 &20.15 &18.99 &18.12 &17.62 \\
\textbf{Uharte-Garazi} &198 &20.46 &17.09 &18.67 &17.82 \\
\textbf{Maule-Lextarre} &198 &11.33 &11.35 &10.41 &10.58 \\
\bottomrule
\end{tabular}
}
\caption{BLEU score of different Basque dialects on all sentences.}
\label{tab:basque-dialects_BLEU}
\end{table*}

\begin{table*}[t!]
\centering
\resizebox{0.9\textwidth}{!}{%
\begin{tabular}{l|rrrr|}\toprule
\multirow{2}{*}{\textbf{Basque}} &\multicolumn{4}{c}{\textbf{BLEU}} \\\cmidrule{2-5}
&\textbf{NLLB-Dis-600M} &\textbf{NLLB-Dis-1.3B} &\textbf{NLLB-1.3B} &\textbf{NLLB-3.3B} \\\midrule
\textbf{Luhuso} &21.38 &19.68 &20.68 &19.54 \\
\textbf{Jutsi} &21.06 &19.35 &19.99 &19.88 \\
\textbf{Muskildi} &9.72 &8.23 &9.41 &8.53 \\
\textbf{Donibane-Lohizune} &20.30 &18.95 &17.93 &17.71 \\
\textbf{Uharte-Garazi} &20.66 &17.16 &18.75 &18.00 \\
\textbf{Maule-Lextarre} &11.29 &11.51 &10.21 &10.34 \\
\bottomrule
\end{tabular}
}
\caption{Comparable BLEU score of different Basque dialects}
\label{tab:basque-dialects-comp_BLEU}
\end{table*}
\label{sec:appendix-italaian}
\begin{table*}[t!]
\centering
\resizebox{\textwidth}{!}{%
\begin{tabular}{l|r|rrrrr|}\toprule
\multirow{2}{*}{\textbf{Italian}} &\multirow{2}{*}{\textbf{\# of Sentences}} &\multicolumn{5}{c}{\textbf{COMET}} \\\cmidrule{3-7}
& &\textbf{DeltaLM} &\textbf{NLLB-Dis-600M} &\textbf{NLLB-Dis-1.3B} &\textbf{NLLB-1.3B} &\textbf{NLLB-3.3B} \\\midrule
\textbf{Albosaggia} &268 &0.6218 &0.6954 &0.7058 &0.7132 &0.7209 \\
\textbf{Aldeno} &1448 &0.7473 &0.8199 &0.8426 &0.8390 &0.8434 \\
\textbf{Altare} &292 &0.5701 &0.6370 &0.6748 &0.6659 &0.6764 \\
\textbf{Arcola} &305 &0.6846 &0.7438 &0.7672 &0.7721 &0.7805 \\
\textbf{Arenzano} &304 &0.6004 &0.6926 &0.7294 &0.7118 &0.7239 \\
\textbf{Ne} &286 &0.6130 &0.7384 &0.7704 &0.7489 &0.7733 \\
\end{tabular}
}
\end{table*}

\begin{table*}[t!]
\centering
\resizebox{\textwidth}{!}{%
\begin{tabular}{l|r|rrrrr|}\toprule
\multirow{2}{*}{\textbf{Italian}} &\multirow{2}{*}{\textbf{\# of Sentences}} &\multicolumn{5}{c}{\textbf{COMET}} \\\cmidrule{3-7}
& &\textbf{DeltaLM} &\textbf{NLLB-Dis-600M} &\textbf{NLLB-Dis-1.3B} &\textbf{NLLB-1.3B} &\textbf{NLLB-3.3B} \\\midrule
\textbf{Treviso} &107 &0.7397 &0.8254 &0.8629 &0.8476 &0.8517 \\
\textbf{Montecchio Maggiore} &127 &0.7650 &0.8364 &0.8633 &0.8576 &0.8567 \\
\textbf{Amblar} &127 &0.6629 &0.7417 &0.7620 &0.7638 &0.7687 \\
\textbf{Andreis} &127 &0.6368 &0.7156 &0.7507 &0.7189 &0.7439 \\
\textbf{Aquileia} &198 &0.6151 &0.7236 &0.7421 &0.7437 &0.7457 \\
\textbf{Arsiero} &184 &0.7514 &0.8455 &0.8704 &0.8675 &0.8697 \\
\end{tabular}
}
\end{table*}

\begin{table*}[t!]
\centering
\resizebox{\textwidth}{!}{%
\begin{tabular}{l|r|rrrrr|}\toprule
\multirow{2}{*}{\textbf{Italian}} &\multirow{2}{*}{\textbf{\# of Sentences}} &\multicolumn{5}{c}{\textbf{COMET}} \\\cmidrule{3-7}
& &\textbf{DeltaLM} &\textbf{NLLB-Dis-600M} &\textbf{NLLB-Dis-1.3B} &\textbf{NLLB-1.3B} &\textbf{NLLB-3.3B} \\\midrule
\textbf{Luzzara} &127 &0.6210 &0.6771 &0.6869 &0.6821 &0.7054 \\
\textbf{Marostica} &326 &0.7271 &0.8047 &0.8283 &0.8239 &0.8247 \\
\textbf{Maserà di Padova} &127 &0.7527 &0.8239 &0.8394 &0.8464 &0.8471 \\
\textbf{Mason Vicentino} &199 &0.7272 &0.8074 &0.8441 &0.8331 &0.8311 \\
\textbf{Arsiè} &308 &0.7072 &0.7742 &0.8055 &0.8042 &0.8105 \\
\textbf{Mirano} &853 &0.7695 &0.8380 &0.8589 &0.8529 &0.8549 \\
\end{tabular}
}
\end{table*}

\begin{table*}[t!]
\centering
\resizebox{\textwidth}{!}{%
\begin{tabular}{l|r|rrrrr|}\toprule
\multirow{2}{*}{\textbf{Italian}} &\multirow{2}{*}{\textbf{\# of Sentences}} &\multicolumn{5}{c}{\textbf{COMET}} \\\cmidrule{3-7}
& &\textbf{DeltaLM} &\textbf{NLLB-Dis-600M} &\textbf{NLLB-Dis-1.3B} &\textbf{NLLB-1.3B} &\textbf{NLLB-3.3B} \\\midrule
\textbf{Correzzola} &122 &0.7767 &0.8450 &0.8570 &0.8594 &0.8692 \\
\textbf{Agugliaro} &11 &0.7494 &0.8134 &0.8253 &0.8239 &0.8457 \\
\textbf{Vittorio Veneto} &56 &0.7933 &0.8322 &0.8561 &0.8640 &0.8768 \\
\textbf{Ariano Irpino} &218 &0.6570 &0.7970 &0.8180 &0.8154 &0.8051 \\
\textbf{Avellino} &1088 &0.6058 &0.7226 &0.7509 &0.7293 &0.7375 \\
\textbf{Bari} &107 &0.6520 &0.7072 &0.7322 &0.7242 &0.7321 \\
\end{tabular}
}
\end{table*}

\begin{table*}[t!]
\centering
\resizebox{\textwidth}{!}{%
\begin{tabular}{l|r|rrrrr|}\toprule
\multirow{2}{*}{\textbf{Italian}} &\multirow{2}{*}{\textbf{\# of Sentences}} &\multicolumn{5}{c}{\textbf{COMET}} \\\cmidrule{3-7}
& &\textbf{DeltaLM} &\textbf{NLLB-Dis-600M} &\textbf{NLLB-Dis-1.3B} &\textbf{NLLB-1.3B} &\textbf{NLLB-3.3B} \\\midrule
\textbf{Lubriano} &96 &0.7452 &0.7883 &0.8033 &0.7904 &0.7980 \\
\textbf{Maglie} &102 &0.7212 &0.7843 &0.8233 &0.8071 &0.7984 \\
\textbf{Civitanova Marche} &95 &0.8129 &0.8387 &0.8424 &0.8372 &0.8506 \\
\textbf{Martina Franca} &103 &0.5450 &0.6082 &0.6240 &0.6116 &0.6123 \\
\textbf{Trieste} &637 &0.7718 &0.8510 &0.8703 &0.8578 &0.8689 \\
\textbf{Trissino} &234 &0.7560 &0.8370 &0.8696 &0.8661 &0.8593 \\
\bottomrule
\end{tabular}
}
\caption{COMET score of different Italian communes on all sentences.}
\label{tab:italian-communes}
\end{table*}

\begin{table*}[t!]
\centering
\resizebox{0.9\textwidth}{!}{%
\begin{tabular}{l|rrrrr|}\toprule
\multirow{2}{*}{\textbf{Itlaian}} &\multicolumn{5}{c}{\textbf{COMET}} \\\cmidrule{2-6}
&\textbf{DeltaLM} &\textbf{NLLB-Dis-600M} &\textbf{NLLB-Dis-1.3B} &\textbf{NLLB-1.3B} &\textbf{NLLB-3.3B} \\\midrule
\textbf{Albosaggia} &0.6226 &0.6966 &0.7068 &0.7138 &0.7234 \\
\textbf{Aldeno} &0.7480 &0.8190 &0.8422 &0.8383 &0.8439 \\
\textbf{Altare} &0.5717 &0.6393 &0.6755 &0.6650 &0.6778 \\
\textbf{Arcola} &0.6846 &0.7449 &0.7659 &0.7734 &0.7796 \\
\textbf{Arenzano} &0.6039 &0.6936 &0.7280 &0.7128 &0.7239 \\
\textbf{Ne} &0.6119 &0.7339 &0.7709 &0.7446 &0.7691 \\
\end{tabular}
}
\end{table*}

\begin{table*}[t!]
\centering
\resizebox{0.9\textwidth}{!}{%
\begin{tabular}{l|rrrrr|}\toprule
\multirow{2}{*}{\textbf{Itlaian}} &\multicolumn{5}{c}{\textbf{COMET}} \\\cmidrule{2-6}
&\textbf{DeltaLM} &\textbf{NLLB-Dis-600M} &\textbf{NLLB-Dis-1.3B} &\textbf{NLLB-1.3B} &\textbf{NLLB-3.3B} \\\midrule
\textbf{Rovereto} &0.7717 &0.8507 &0.8739 &0.8725 &0.8572 \\
\textbf{Salzano} &0.7228 &0.8309 &0.8510 &0.8483 &0.8495 \\
\textbf{San Michele al Tagliamento} &0.6534 &0.7436 &0.7621 &0.7584 &0.7616 \\
\textbf{Scorzè} &0.7609 &0.8233 &0.8615 &0.8583 &0.8530 \\
\textbf{Selva di Val Gardena} &0.5664 &0.6448 &0.6731 &0.6686 &0.6652 \\
\textbf{Tezze sul Brenta} &0.7400 &0.8240 &0.8440 &0.8394 &0.8364 \\
\end{tabular}
}
\end{table*}

\begin{table*}[t!]
\centering
\resizebox{0.9\textwidth}{!}{%
\begin{tabular}{l|rrrrr|}\toprule
\multirow{2}{*}{\textbf{Itlaian}} &\multicolumn{5}{c}{\textbf{COMET}} \\\cmidrule{2-6}
&\textbf{DeltaLM} &\textbf{NLLB-Dis-600M} &\textbf{NLLB-Dis-1.3B} &\textbf{NLLB-1.3B} &\textbf{NLLB-3.3B} \\\midrule
\textbf{Cittadella} &0.7455 &0.8175 &0.8455 &0.8408 &0.8422 \\
\textbf{Falcade} &0.6657 &0.7095 &0.7326 &0.7264 &0.7342 \\
\textbf{Sernaglia della Battaglia} &0.7268 &0.7978 &0.8102 &0.8064 &0.8285 \\
\textbf{Ferrara} &0.6116 &0.7036 &0.7163 &0.7194 &0.7190 \\
\textbf{Sondalo} &0.6281 &0.7172 &0.7390 &0.7525 &0.7412 \\
\textbf{Galliera Veneta} &0.7470 &0.8158 &0.8367 &0.8318 &0.8396 \\
\end{tabular}
}
\end{table*}

\begin{table*}[t!]
\centering
\resizebox{0.9\textwidth}{!}{%
\begin{tabular}{l|rrrrr|}\toprule
\multirow{2}{*}{\textbf{Itlaian}} &\multicolumn{5}{c}{\textbf{COMET}} \\\cmidrule{2-6}
&\textbf{DeltaLM} &\textbf{NLLB-Dis-600M} &\textbf{NLLB-Dis-1.3B} &\textbf{NLLB-1.3B} &\textbf{NLLB-3.3B} \\\midrule
\textbf{Selvazzano Dentro} &0.7490 &0.8178 &0.8426 &0.8465 &0.8331 \\
\textbf{Valdidentro} &0.6587 &0.7375 &0.7555 &0.7494 &0.7488 \\
\textbf{Solesino} &0.7757 &0.8358 &0.8600 &0.8503 &0.8367 \\
\textbf{Calasetta} &0.5142 &0.6494 &0.6897 &0.6862 &0.6756 \\
\textbf{Taggia} &0.7094 &0.7870 &0.8093 &0.8023 &0.8109 \\
\textbf{Taglio di Po} &0.6965 &0.7822 &0.7858 &0.7836 &0.7909 \\
\end{tabular}
}
\end{table*}

\begin{table*}[t!]
\centering
\resizebox{0.9\textwidth}{!}{%
\begin{tabular}{l|rrrrr|}\toprule
\multirow{2}{*}{\textbf{Itlaian}} &\multicolumn{5}{c}{\textbf{COMET}} \\\cmidrule{2-6}
&\textbf{DeltaLM} &\textbf{NLLB-Dis-600M} &\textbf{NLLB-Dis-1.3B} &\textbf{NLLB-1.3B} &\textbf{NLLB-3.3B} \\\midrule
\textbf{San Cesario di Lecce} &0.7481 &0.8000 &0.8274 &0.8143 &0.8181 \\
\textbf{San Marco in Lamis} &0.7022 &0.7617 &0.7746 &0.7848 &0.7788 \\
\textbf{San Martino in Pensilis} &0.4193 &0.6121 &0.6844 &0.6908 &0.7033 \\
\textbf{Sciacca} &0.7333 &0.7744 &0.7986 &0.7775 &0.7911 \\
\textbf{Terravecchia} &0.5993 &0.7373 &0.7617 &0.7517 &0.7633 \\
\textbf{Trepuzzi} &0.6663 &0.7262 &0.7512 &0.7376 &0.7365 \\
\bottomrule
\end{tabular}
}
\caption{Comparable COMET score of different Italian communes.}
\label{tab:italian-communes-comp}
\end{table*}
\begin{table*}[t!]
\centering
\resizebox{\textwidth}{!}{%
\begin{tabular}{l|r|rrrrr|}\toprule
\multirow{2}{*}{\textbf{Italian}} &\multirow{2}{*}{\textbf{\# of Sentences}} &\multicolumn{5}{c}{\textbf{BLEU}} \\\cmidrule{3-7}
& &\textbf{DeltaLM} &\textbf{NLLB-Dis-600M} &\textbf{NLLB-Dis-1.3B} &\textbf{NLLB-1.3B} &\textbf{NLLB-3.3B} \\\midrule
\textbf{Albosaggia} &268 &1.47 &14.78 &15.00 &15.35 &14.53 \\
\textbf{Aldeno} &1448 &9.72 &27.33 &32.14 &30.51 &32.16 \\
\textbf{Altare} &292 &2.02 &9.57 &12.63 &10.66 &11.70 \\
\textbf{Arcola} &305 &4.66 &16.23 &17.89 &18.32 &19.48 \\
\textbf{Arenzano} &304 &2.77 &13.00 &16.61 &13.79 &15.40 \\
\textbf{Ne} &286 &1.90 &17.78 &21.40 &17.13 &21.19 \\
\end{tabular}
}
\end{table*}

\begin{table*}[t!]
\centering
\resizebox{\textwidth}{!}{%
\begin{tabular}{l|r|rrrrr|}\toprule
\multirow{2}{*}{\textbf{Italian}} &\multirow{2}{*}{\textbf{\# of Sentences}} &\multicolumn{5}{c}{\textbf{BLEU}} \\\cmidrule{3-7}
& &\textbf{DeltaLM} &\textbf{NLLB-Dis-600M} &\textbf{NLLB-Dis-1.3B} &\textbf{NLLB-1.3B} &\textbf{NLLB-3.3B} \\\midrule
\textbf{Treviso} &107 &7.37 &34.86 &43.43 &35.07 &36.43 \\
\textbf{Trieste} &637 &12.45 &34.52 &38.30 &35.43 &37.17 \\
\textbf{Trissino} &234 &8.21 &33.47 &40.50 &36.21 &38.49 \\
\textbf{Vallecrosia} &304 &4.22 &16.78 &21.57 &18.98 &20.84 \\
\textbf{Vaprio d'Adda} &220 &1.62 &14.62 &12.77 &14.48 &14.59 \\
\textbf{Vione} &107 &4.12 &11.06 &13.80 &16.96 &15.48 \\
\end{tabular}
}
\end{table*}

\begin{table*}[t!]
\centering
\resizebox{\textwidth}{!}{%
\begin{tabular}{l|r|rrrrr|}\toprule
\multirow{2}{*}{\textbf{Italian}} &\multirow{2}{*}{\textbf{\# of Sentences}} &\multicolumn{5}{c}{\textbf{BLEU}} \\\cmidrule{3-7}
& &\textbf{DeltaLM} &\textbf{NLLB-Dis-600M} &\textbf{NLLB-Dis-1.3B} &\textbf{NLLB-1.3B} &\textbf{NLLB-3.3B} \\\midrule
\textbf{Luzzara} &127 &3.21 &13.07 &14.04 &12.58 &14.41 \\
\textbf{Marostica} &326 &8.45 &27.62 &30.88 &28.79 &32.56 \\
\textbf{Maserà di Padova} &127 &9.16 &28.80 &33.82 &30.18 &33.93 \\
\textbf{Mason Vicentino} &199 &9.61 &26.54 &31.90 &28.29 &32.07 \\
\textbf{Arsiè} &308 &5.38 &19.74 &25.80 &23.11 &25.80 \\
\textbf{Mirano} &853 &11.47 &31.99 &34.96 &32.56 &35.74 \\
\end{tabular}
}
\end{table*}

\begin{table*}[t!]
\centering
\resizebox{\textwidth}{!}{%
\begin{tabular}{l|r|rrrrr|}\toprule
\multirow{2}{*}{\textbf{Italian}} &\multirow{2}{*}{\textbf{\# of Sentences}} &\multicolumn{5}{c}{\textbf{BLEU}} \\\cmidrule{3-7}
& &\textbf{DeltaLM} &\textbf{NLLB-Dis-600M} &\textbf{NLLB-Dis-1.3B} &\textbf{NLLB-1.3B} &\textbf{NLLB-3.3B} \\\midrule
\textbf{Correzzola} &122 &13.31 &35.29 &37.33 &34.02 &40.72 \\
\textbf{Agugliaro} &11 &6.38 &31.50 &27.29 &28.26 &34.44 \\
\textbf{Vittorio Veneto} &56 &17.63 &19.69 &23.68 &26.45 &33.33 \\
\textbf{Ariano Irpino} &218 &4.16 &26.30 &27.74 &24.31 &23.98 \\
\textbf{Avellino} &1088 &2.50 &15.37 &17.00 &14.99 &15.16 \\
\textbf{Bari} &107 &0.74 &10.94 &14.95 &13.11 &13.16 \\
\end{tabular}
}
\end{table*}

\begin{table*}[t!]
\centering
\resizebox{\textwidth}{!}{%
\begin{tabular}{l|r|rrrrr|}\toprule
\multirow{2}{*}{\textbf{Italian}} &\multirow{2}{*}{\textbf{\# of Sentences}} &\multicolumn{5}{c}{\textbf{BLEU}} \\\cmidrule{3-7}
& &\textbf{DeltaLM} &\textbf{NLLB-Dis-600M} &\textbf{NLLB-Dis-1.3B} &\textbf{NLLB-1.3B} &\textbf{NLLB-3.3B} \\\midrule
\textbf{Lubriano} &96 &7.61 &17.83 &18.98 &15.65 &19.96 \\
\textbf{Maglie} &102 &5.04 &21.68 &27.33 &24.29 &25.41 \\
\textbf{Civitanova Marche} &95 &14.67 &26.31 &25.99 &23.76 &26.08 \\
\textbf{Martina Franca} &103 &0.37 &4.39 &5.91 &5.09 &5.16 \\
\textbf{Martinsicuro} &101 &0.99 &8.19 &11.38 &10.71 &8.81 \\
\textbf{Massafra} &104 &2.39 &9.29 &9.10 &11.54 &8.99 \\
\bottomrule
\end{tabular}
}
\caption{BLEU score of different Italian communes on all sentences.}
\label{tab:italian-communes_BLEU}
\end{table*}

\begin{table*}[t!]
\centering
\resizebox{0.9\textwidth}{!}{%
\begin{tabular}{l|rrrrr|}\toprule
\multirow{2}{*}{\textbf{Itlaian}} &\multicolumn{5}{c}{\textbf{BLEU}} \\\cmidrule{2-6}
&\textbf{DeltaLM} &\textbf{NLLB-Dis-600M} &\textbf{NLLB-Dis-1.3B} &\textbf{NLLB-1.3B} &\textbf{NLLB-3.3B} \\\midrule
\textbf{Albosaggia} &8.33 &14.62 &14.98 &15.63 &14.68 \\
\textbf{Aldeno} &18.31 &26.83 &31.52 &30.30 &32.03 \\
\textbf{Altare} &8.00 &9.73 &12.50 &10.72 &11.77 \\
\textbf{Arcola} &12.60 &16.33 &18.11 &18.56 &19.71 \\
\textbf{Arenzano} &8.32 &13.17 &16.55 &14.23 &15.45 \\
\textbf{Ne} &8.31 &16.90 &20.67 &16.59 &20.38 \\
\end{tabular}
}
\end{table*}

\begin{table*}[t!]
\centering
\resizebox{0.9\textwidth}{!}{%
\begin{tabular}{l|rrrrr|}\toprule
\multirow{2}{*}{\textbf{Itlaian}} &\multicolumn{5}{c}{\textbf{BLEU}} \\\cmidrule{2-6}
&\textbf{DeltaLM} &\textbf{NLLB-Dis-600M} &\textbf{NLLB-Dis-1.3B} &\textbf{NLLB-1.3B} &\textbf{NLLB-3.3B} \\\midrule
\textbf{Rovereto} &23.56 &35.34 &41.92 &42.71 &39.37 \\
\textbf{Salzano} &16.19 &29.93 &35.28 &32.41 &36.39 \\
\textbf{San Michele al Tagliamento} &11.63 &17.69 &20.96 &19.85 &20.99 \\
\textbf{Scorzè} &21.32 &31.45 &34.98 &34.61 &34.09 \\
\textbf{Selva di Val Gardena} &7.71 &10.69 &11.95 &11.59 &12.35 \\
\textbf{Tezze sul Brenta} &17.76 &29.63 &34.60 &30.73 &32.77 \\
\end{tabular}
}
\end{table*}

\begin{table*}[t!]
\centering
\resizebox{0.9\textwidth}{!}{%
\begin{tabular}{l|rrrrr|}\toprule
\multirow{2}{*}{\textbf{Itlaian}} &\multicolumn{5}{c}{\textbf{BLEU}} \\\cmidrule{2-6}
&\textbf{DeltaLM} &\textbf{NLLB-Dis-600M} &\textbf{NLLB-Dis-1.3B} &\textbf{NLLB-1.3B} &\textbf{NLLB-3.3B} \\\midrule
\textbf{Cittadella} &18.10 &30.28 &34.98 &31.50 &35.46 \\
\textbf{Falcade} &10.96 &11.98 &14.47 &13.40 &17.07 \\
\textbf{Sernaglia della Battaglia} &16.39 &24.28 &29.58 &27.34 &33.45 \\
\textbf{Ferrara} &9.21 &13.54 &15.72 &13.86 &15.59 \\
\textbf{Sondalo} &8.45 &15.90 &17.60 &18.36 &19.04 \\
\textbf{Galliera Veneta} &18.79 &30.50 &34.50 &30.30 &35.71 \\
\end{tabular}
}
\end{table*}

\begin{table*}[t!]
\centering
\resizebox{0.9\textwidth}{!}{%
\begin{tabular}{l|rrrrr|}\toprule
\multirow{2}{*}{\textbf{Itlaian}} &\multicolumn{5}{c}{\textbf{BLEU}} \\\cmidrule{2-6}
&\textbf{DeltaLM} &\textbf{NLLB-Dis-600M} &\textbf{NLLB-Dis-1.3B} &\textbf{NLLB-1.3B} &\textbf{NLLB-3.3B} \\\midrule
\textbf{Selvazzano Dentro} &18.32 &28.95 &34.80 &31.11 &34.93 \\
\textbf{Valdidentro} &11.35 &15.02 &17.67 &15.56 &18.05 \\
\textbf{Solesino} &22.05 &28.29 &38.45 &33.73 &33.50 \\
\textbf{Calasetta} &5.34 &8.53 &10.27 &10.51 &9.26 \\
\textbf{Taggia} &19.21 &27.82 &31.81 &28.56 &30.19 \\
\textbf{Taglio di Po} &13.17 &19.45 &21.09 &19.85 &22.72 \\
\end{tabular}
}
\end{table*}

\begin{table*}[t!]
\centering
\resizebox{0.9\textwidth}{!}{%
\begin{tabular}{l|rrrrr|}\toprule
\multirow{2}{*}{\textbf{Itlaian}} &\multicolumn{5}{c}{\textbf{BLEU}} \\\cmidrule{2-6}
&\textbf{DeltaLM} &\textbf{NLLB-Dis-600M} &\textbf{NLLB-Dis-1.3B} &\textbf{NLLB-1.3B} &\textbf{NLLB-3.3B} \\\midrule
\textbf{San Cesario di Lecce} &20.69 &28.06 &30.93 &29.72 &27.80 \\
\textbf{San Marco in Lamis} &14.11 &20.38 &21.89 &20.46 &20.96 \\
\textbf{San Martino in Pensilis} &2.21 &7.90 &14.49 &12.41 &14.54 \\
\textbf{Sciacca} &17.02 &27.92 &24.31 &23.58 &21.45 \\
\textbf{Terravecchia} &10.91 &14.31 &17.20 &14.47 &16.57 \\
\textbf{Trepuzzi} &10.90 &18.83 &19.11 &17.29 &19.46 \\
\bottomrule
\end{tabular}
}
\caption{Comparable BLEU score of different Italian communes.}
\label{tab:italian-communes-comp_BLEU}
\end{table*}
\begin{table*}[t!]
\centering
\resizebox{\textwidth}{!}{%
\begin{tabular}{l|r|rrrrr|}\toprule
\multirow{2}{*}{\textbf{Italian}} &\multirow{2}{*}{\textbf{\# of Sentences}} &\multicolumn{5}{c}{\textbf{COMET}} \\\cmidrule{3-7}
& &\textbf{DeltaLM} &\textbf{NLLB-Dis-600M} &\textbf{NLLB-Dis-1.3B} &\textbf{NLLB-1.3B} &\textbf{NLLB-3.3B} \\\midrule
\textbf{Lombardia} &8027 &0.6209 &0.7091 &0.7319 &0.7281 &0.7342 \\
\textbf{Trentino Alto Adige} &3787 &0.6871 &0.7637 &0.7859 &0.7845 &0.7834 \\
\textbf{Liguria} &5939 &0.6404 &0.7277 &0.7588 &0.7467 &0.7578 \\
\textbf{Veneto} &21723 &0.7330 &0.8066 & 0.8280 &0.8234 &0.8255 \\
\textbf{Emilia Romagna} &2125 &0.6028 &0.6854 &0.7071 &0.6997 &0.7091 \\
\textbf{Piemonte} &4264 &0.6048 &0.6914 &0.7179 &0.7074 &0.7166 \\
\textbf{Friuli Venezia Giulia} &3878 &0.6526 &0.7439 &0.7675 &0.7598 &0.7760 \\
\textbf{Toscana} &1047 &0.7452 &0.7943 & 0.8116 &0.8086 &0.8174 \\
\textbf{Sicilia} &5500 &0.6700 &0.7752 &0.7941 &0.7849 &0.7857 \\
\textbf{Marche} &717 &0.7140 &0.7775 &0.7977 &0.7923 &0.7984 \\
\textbf{Sardegna} &1065 &0.5778 &0.6779 &0.7080 &0.6987 &0.7031 \\
\textbf{Puglia} &6100 &0.6470 &0.7236 &0.7490 &0.7343 &0.7401 \\
\textbf{Campania} &2901 &0.6083 &0.7342 &0.7614 &0.7483 &0.7562 \\
\textbf{Calabria} &1321 &0.6469 &0.7612 &0.7883 &0.7746 &0.7774 \\
\textbf{Basilicata} &326 &0.5502 &0.6992 &0.7299 &0.7315 &0.7166 \\
\textbf{Umbria} &303 &0.8373 &0.8650 & 0.8766 &0.8654 &0.8748 \\
\textbf{Abruzzo} &1785 &0.5633 &0.6920 &0.6896 &0.6931 &0.6997 \\
\textbf{Molise} &229 &0.6059 &0.7101 &0.7431 &0.7205 &0.7359 \\
\textbf{Lazio} &526 &0.8007 &0.8324 & 0.8417 &0.8386 &0.8509 \\
\bottomrule
\end{tabular}
}
\caption{COMET score of different Italian regions on all sentences.}
\label{tab:italian-regions}
\end{table*}

\begin{table*}[t!]
\centering
\resizebox{0.9\textwidth}{!}{%
\begin{tabular}{l|rrrrrr|}\toprule
\multirow{2}{*}{\textbf{Italian}} &\multicolumn{5}{c}{\textbf{COMET}} \\\cmidrule{2-6}
&\textbf{DeltaLM} &\textbf{NLLB-Dis-600M} &\textbf{NLLB-Dis-1.3B} &\textbf{NLLB-1.3B} &\textbf{NLLB-3.3B} \\\midrule
\textbf{Lombardia} &0.6257 &0.7103 &0.7316 &0.7278 &0.7341 \\
\textbf{Trentino Alto Adige} &0.6826 &0.7584 &0.7793 &0.7805 &0.7763 \\
\textbf{Liguria} &0.6445 &0.7311 &0.7612 &0.7495 &0.7604 \\
\textbf{Veneto} &0.7400 &0.8117 &\textbf{0.8330} &0.8276 &0.8311 \\
\textbf{Emilia Romagna} &0.6034 &0.6848 &0.7071 &0.6981 &0.7109 \\
\textbf{Piemonte} &0.6113 &0.6969 &0.7266 &0.7139 &0.7231 \\
\textbf{Friuli Venezia Giulia} &0.6456 &0.7378 &0.7614 &0.7537 &0.7695 \\
\textbf{Toscana} &0.7272 &0.7815 &\textbf{0.7991} &0.7961 &0.8051 \\
\textbf{Sicilia} &0.6627 &0.7654 &0.7857 &0.7758 &0.7764 \\
\textbf{Marche} &0.7253 &0.7822 &\textbf{0.7996} &0.7951 &0.8016 \\
\textbf{Sardegna} &0.5820 &0.6777 &0.7046 &0.6928 &0.7016 \\
\textbf{Puglia} &0.6507 &0.7241 &0.7493 &0.7323 &0.7396 \\
\textbf{Campania} &0.5821 &0.7235 &0.7545 &0.7420 &0.7511 \\
\textbf{Calabria} &0.6498 &0.7644 &0.7914 &0.7770 &0.7801 \\
\textbf{Basilicata} &0.5322 &0.7067 &0.7451 &0.7419 &0.7296 \\
\textbf{Umbria} &0.8240 &0.8611 &\textbf{0.8720} &0.8594 &0.8689 \\
\textbf{Abruzzo} &0.5622 &0.6915 &0.6880 &0.6915 &0.6990 \\
\textbf{Molise} &0.5833 &0.6968 &0.7372 &0.7191 &0.7339 \\
\textbf{Lazio} &0.8024 &0.8342 &\textbf{0.8423} &0.8406 &0.8529 \\
\bottomrule
\end{tabular}
}
\caption{Comparable COMET score of different Italian regions.}
\label{tab:italian-regions-comp}
\end{table*}

\begin{table*}[t!]
\centering
\resizebox{\textwidth}{!}{%
\begin{tabular}{l|r|rrrrr|}\toprule
\multirow{2}{*}{\textbf{Italian}} &\multirow{2}{*}{\textbf{\# of Sentences}} &\multicolumn{5}{c}{\textbf{BLEU}} \\\cmidrule{3-7}
& &\textbf{DeltaLM} &\textbf{NLLB-Dis-600M} &\textbf{NLLB-Dis-1.3B} &\textbf{NLLB-1.3B} &\textbf{NLLB-3.3B} \\\midrule
\textbf{Lombardia} &8027 &3.04 &15.01 &17.40 &16.23 &16.88 \\
\textbf{Trentino Alto Adige} &3787 &6.58 &20.98 &24.71 &23.53 &24.46 \\
\textbf{Liguria} &5939 &3.92 &17.34 &20.39 &18.66 &20.08 \\
\textbf{Veneto} &21723 &8.36 &26.92 &31.20 &27.97 &31.13 \\
\textbf{Emilia Romagna} &2125 &2.36 &13.22 &15.57 &14.01 &15.07 \\
\textbf{Piemonte} &4264 &2.39 &13.14 &16.17 &14.30 &15.39 \\
\textbf{Friuli Venezia Giulia} &3878 &4.64 &19.03 &22.96 &20.90 &22.84 \\
\textbf{Toscana} &1047 &21.73 &32.67 &36.61 &35.74 &37.51 \\
\textbf{Sicilia} &5500 &4.03 &23.55 &25.11 &23.72 &22.76 \\
\textbf{Marche} &717 &7.50 &22.49 &26.00 &23.76 &27.66 \\
\textbf{Sardegna} &1065 &1.36 &11.23 &13.67 &12.63 &12.75 \\
\textbf{Puglia} &6100 &3.16 &16.28 &17.86 &15.51 &16.84 \\
\textbf{Campania} &2901 &3.63 &16.92 &18.19 &16.53 &17.03 \\
\textbf{Calabria} &1321 &3.94 &19.90 &22.49 &20.67 &20.28 \\
\textbf{Basilicata} &326 &0.41 &11.87 &13.99 &13.13 &12.94 \\
\textbf{Umbria} &303 &23.71 &37.15 &39.83 &36.00 &36.88 \\
\textbf{Abruzzo} &1785 &2.41 &10.08 &10.55 &8.70 &10.86 \\
\textbf{Molise} &229 &3.07 &13.07 &16.12 &11.98 &16.70 \\
\textbf{Lazio} &526 &14.39 &27.27 &28.66 &26.34 &30.14 \\
\bottomrule
\end{tabular}
}
\caption{BLEU score of different Italian regions on all sentences.}
\label{tab:italian-regions_BLEU}
\end{table*}

\begin{table*}[t!]
\centering
\resizebox{0.9\textwidth}{!}{%
\begin{tabular}{l|rrrrrr|}\toprule
\multirow{2}{*}{\textbf{Italian}} &\multicolumn{5}{c}{\textbf{BLEU}} \\\cmidrule{2-6}
&\textbf{DeltaLM} &\textbf{NLLB-Dis-600M} &\textbf{NLLB-Dis-1.3B} &\textbf{NLLB-1.3B} &\textbf{NLLB-3.3B} \\\midrule
\textbf{Lombardia} &10.02 &15.20 &17.66 &16.69 &17.12 \\
\textbf{Trentino Alto Adige} &13.99 &20.27 &23.97 &23.24 &23.70 \\
\textbf{Liguria} &11.53 &17.90 &20.83 &19.09 &20.49 \\
\textbf{Veneto} &17.94 &27.68 &31.92 &28.62 &32.01 \\
\textbf{Emilia Romagna} &9.13 &12.98 &15.32 &13.66 &15.04 \\
\textbf{Piemonte} &9.04 &13.65 &16.99 &14.94 &15.89 \\
\textbf{Friuli Venezia Giulia} &11.39 &18.12 &22.23 &20.21 &21.88 \\
\textbf{Toscana} &26.36 &30.32 &34.15 &33.44 &34.98 \\
\textbf{Sicilia} &11.62 &22.12 &23.78 &22.28 &21.56 \\
\textbf{Marche} &17.17 &22.79 &26.14 &24.12 &27.90 \\
\textbf{Sardegna} &7.09 &11.14 &13.15 &12.17 &12.15 \\
\textbf{Puglia} &10.50 &15.86 &17.71 &15.02 &16.65 \\
\textbf{Campania} &7.45 &15.68 &16.85 &15.49 &16.10 \\
\textbf{Calabria} &12.16 &20.06 &22.53 &20.98 &20.78 \\
\textbf{Basilicata} &5.89 &12.80 &15.52 &13.79 &14.00 \\
\textbf{Umbria} &30.18 &35.05 &37.09 &32.66 &34.58 \\
\textbf{Abruzzo} &6.48 &10.15 &10.46 &8.60 &10.72 \\
\textbf{Molise} &8.11 &12.71 &16.18 &12.54 &16.64 \\
\textbf{Lazio} &22.81 &27.95 &28.94 &27.29 &30.43 \\
\bottomrule
\end{tabular}
}
\caption{Comparable BLEU score of different Italian regions.}
\label{tab:italian-regions-comp_BLEU}
\end{table*}
\clearpage
\label{sec:appendix-swiss}
\begin{table*}[t!]
\centering
\resizebox{0.9\textwidth}{!}{%
\begin{tabular}{l|r|rrrrr|}\toprule
\multirow{2}{*}{\textbf{Swiss-German}} &\multirow{2}{*}{\textbf{\# of Sentences}} &\multicolumn{4}{c}{\textbf{COMET}} \\\cmidrule{3-6}
& &\textbf{NLLB-Dis-600M} &\textbf{NLLB-Dis-1.3B} &\textbf{NLLB-1.3B} &\textbf{NLLB-3.3B} \\\midrule
\textbf{Aarau,AG} &121 &0.8734 &0.8787 &0.8714 &0.8882 \\
\textbf{Aarberg,BE} &117 &0.8701 &0.8772 &0.8616 &0.8839 \\
\textbf{Aarburg,AG} &118 &0.8706 &0.8808 &0.8663 &0.8905 \\
\textbf{Adelboden,BE} &120 &0.8686 &0.8684 &0.8675 &0.8829 \\
\textbf{Aedermannsdorf,SO} &115 &0.8655 &0.8744 &0.8591 &0.8806 \\
\textbf{Aesch,BL} &118 &0.8712 &0.8759 &0.8688 &0.8865 \\
\end{tabular}
}
\end{table*}

\begin{table*}[t!]
\centering
\resizebox{0.9\textwidth}{!}{%
\begin{tabular}{l|r|rrrrr|}\toprule
\multirow{2}{*}{\textbf{Swiss-German}} &\multirow{2}{*}{\textbf{\# of Sentences}} &\multicolumn{4}{c}{\textbf{COMET}} \\\cmidrule{3-6}
& &\textbf{NLLB-Dis-600M} &\textbf{NLLB-Dis-1.3B} &\textbf{NLLB-1.3B} &\textbf{NLLB-3.3B} \\\midrule
\textbf{Densbüren,AG} &121 &0.8732 &0.8762 &0.8704 &0.8866 \\
\textbf{Diemtigen,BE} &118 &0.8676 &0.8775 &0.8674 &0.8850 \\
\textbf{Diepoldsau,SG} &113 &0.8732 &0.8849 &0.8719 &0.8898 \\
\textbf{Diessbach bei Büren,BE} &115 &0.8657 &0.8771 &0.8635 &0.8867 \\
\textbf{Düdingen,FR} &114 &0.8679 &0.8765 &0.8633 &0.8881 \\
\textbf{Ebnat-Kappel,SG} &122 &0.8757 &0.8783 &0.8738 &0.8873 \\
\end{tabular}
}
\end{table*}

\begin{table*}[t!]
\centering
\resizebox{0.9\textwidth}{!}{%
\begin{tabular}{l|r|rrrrr|}\toprule
\multirow{2}{*}{\textbf{Swiss-German}} &\multirow{2}{*}{\textbf{\# of Sentences}} &\multicolumn{4}{c}{\textbf{COMET}} \\\cmidrule{3-6}
& &\textbf{NLLB-Dis-600M} &\textbf{NLLB-Dis-1.3B} &\textbf{NLLB-1.3B} &\textbf{NLLB-3.3B} \\\midrule
\textbf{Schlatt-Haslen,AI} &112 &0.8656 &0.8806 &0.8685 &0.8847 \\
\textbf{Hedingen,ZH} &116 &0.8710 &0.8821 &0.8660 &0.8862 \\
\textbf{Heiden,AR} &118 &0.8707 &0.8825 &0.8724 &0.8909 \\
\textbf{Heitenried,FR} &118 &0.8622 &0.8710 &0.8538 &0.8740 \\
\textbf{Herisau,AR} &113 &0.8729 &0.8826 &0.8731 &0.8894 \\
\textbf{Hölstein,BL} &120 &0.8711 &0.8735 &0.8644 &0.8858 \\
\end{tabular}
}
\end{table*}

\begin{table*}[t!]
\centering
\resizebox{0.9\textwidth}{!}{%
\begin{tabular}{l|r|rrrrr|}\toprule
\multirow{2}{*}{\textbf{Swiss-German}} &\multirow{2}{*}{\textbf{\# of Sentences}} &\multicolumn{4}{c}{\textbf{COMET}} \\\cmidrule{3-6}
& &\textbf{NLLB-Dis-600M} &\textbf{NLLB-Dis-1.3B} &\textbf{NLLB-1.3B} &\textbf{NLLB-3.3B} \\\midrule
\textbf{Magden,AG} &114 &0.8729 &0.8739 &0.8663 &0.8852 \\
\textbf{Maisprach,BL} &116 &0.8705 &0.8725 &0.8666 &0.8836 \\
\textbf{Malans,GR} &114 &0.8772 &0.8802 &0.8750 &0.8879 \\
\textbf{Malters,LU} &117 &0.8711 &0.8729 &0.8664 &0.8856 \\
\textbf{Mammern,TG} &120 &0.8776 &0.8821 &0.8738 &0.8881 \\
\textbf{Marbach,LU} &121 &0.8769 &0.8793 &0.8732 &0.8899 \\
\end{tabular}
}
\end{table*}

\begin{table*}[t!]
\centering
\resizebox{0.9\textwidth}{!}{%
\begin{tabular}{l|r|rrrrr|}\toprule
\multirow{2}{*}{\textbf{Swiss-German}} &\multirow{2}{*}{\textbf{\# of Sentences}} &\multicolumn{4}{c}{\textbf{COMET}} \\\cmidrule{3-6}
& &\textbf{NLLB-Dis-600M} &\textbf{NLLB-Dis-1.3B} &\textbf{NLLB-1.3B} &\textbf{NLLB-3.3B} \\\midrule
\textbf{Reutigen,BE} &118 &0.8652 &0.8781 &0.8688 &0.8844 \\
\textbf{Rheineck,SG} &119 &0.8695 &0.8823 &0.8670 &0.8877 \\
\textbf{Medels im Rheinwald,GR} &111 &0.8760 &0.8773 &0.8668 &0.8843 \\
\textbf{Wattwil,SG} &117 &0.8700 &0.8826 &0.8668 &0.8866 \\
\textbf{Rickenbach,SO} &118 &0.8697 &0.8733 &0.8681 &0.8843 \\
\textbf{Rifferswil,ZH} &114 &0.8731 &0.8864 &0.8694 &0.8927 \\
\end{tabular}
}
\end{table*}

\begin{table*}[t!]
\centering
\resizebox{0.9\textwidth}{!}{%
\begin{tabular}{l|r|rrrrr|}\toprule
\multirow{2}{*}{\textbf{Swiss-German}} &\multirow{2}{*}{\textbf{\# of Sentences}} &\multicolumn{4}{c}{\textbf{COMET}} \\\cmidrule{3-6}
& &\textbf{NLLB-Dis-600M} &\textbf{NLLB-Dis-1.3B} &\textbf{NLLB-1.3B} &\textbf{NLLB-3.3B} \\\midrule
\textbf{Triengen,LU} &118 &0.8692 &0.8734 &0.8679 &0.8840 \\
\textbf{Trimmis,GR} &117 &0.8662 &0.8803 &0.8682 &0.8864 \\
\textbf{Trogen,AR} &118 &0.8692 &0.8825 &0.8693 &0.8870 \\
\textbf{Tüscherz-Alfermée,BE} &115 &0.8706 &0.8761 &0.8696 &0.8865 \\
\textbf{Tuggen,SZ} &120 &0.8787 &0.8833 &0.8741 &0.8920 \\
\textbf{Turbenthal,ZH} &124 &0.8774 &0.8832 &0.8755 &0.8901 \\
\bottomrule
\end{tabular}
}
\caption{COMET score of different Swiss-German dialects on all sentences.}
\label{tab:swiss-dialects}
\end{table*}

\begin{table*}[t!]
\centering
\resizebox{0.8\textwidth}{!}{%
\begin{tabular}{l|rrrr|}\toprule
\multirow{2}{*}{\textbf{Swiss-German}} &\multicolumn{4}{c}{\textbf{COMET}} \\\cmidrule{2-5}
&\textbf{NLLB-Dis-600M} &\textbf{NLLB-Dis-1.3B} &\textbf{NLLB-1.3B} &\textbf{NLLB-3.3B} \\\midrule
\textbf{Aarau,AG} &0.8723 &0.8784 &0.8725 &0.8881 \\
\textbf{Aarberg,BE} &0.8707 &0.8774 &0.8628 &0.8841 \\
\textbf{Aarburg,AG} &0.8697 &0.8805 &0.8655 &0.8900 \\
\textbf{Adelboden,BE} &0.8678 &0.8677 &0.8671 &0.8827 \\
\textbf{Aedermannsdorf,SO} &0.8645 &0.8738 &0.8588 &0.8804 \\
\textbf{Aesch,BL} &0.8703 &0.8752 &0.8691 &0.8856 \\
\end{tabular}
}
\end{table*}

\begin{table*}[t!]
\centering
\resizebox{0.8\textwidth}{!}{%
\begin{tabular}{l|rrrr|}\toprule
\multirow{2}{*}{\textbf{Swiss-German}} &\multicolumn{4}{c}{\textbf{COMET}} \\\cmidrule{2-5}
&\textbf{NLLB-Dis-600M} &\textbf{NLLB-Dis-1.3B} &\textbf{NLLB-1.3B} &\textbf{NLLB-3.3B} \\\midrule
\textbf{Dagmersellen,LU} &0.8701 &0.8766 &0.8697 &0.8840 \\
\textbf{Davos,GR} &0.8742 &0.8837 &0.8683 &0.8912 \\
\textbf{Degersheim,SG} &0.8707 &0.8850 &0.8741 &0.8867 \\
\textbf{Densbüren,AG} &0.8740 &0.8778 &0.8721 &0.8881 \\
\textbf{Diemtigen,BE} &0.8677 &0.8774 &0.8664 &0.8846 \\
\textbf{Diepoldsau,SG} &0.8737 &0.8858 &0.8737 &0.8904 \\
\end{tabular}
}
\end{table*}

\begin{table*}[t!]
\centering
\resizebox{0.8\textwidth}{!}{%
\begin{tabular}{l|rrrr|}\toprule
\multirow{2}{*}{\textbf{Swiss-German}} &\multicolumn{4}{c}{\textbf{COMET}} \\\cmidrule{2-5}
&\textbf{NLLB-Dis-600M} &\textbf{NLLB-Dis-1.3B} &\textbf{NLLB-1.3B} &\textbf{NLLB-3.3B} \\\midrule
\textbf{Gurtnellen,UR} &0.8757 &0.8778 &0.8695 &0.8825 \\
\textbf{Guttannen,BE} &0.8671 &0.8738 &0.8687 &0.8828 \\
\textbf{Guttet-Feschel,VS} &0.8701 &0.8747 &0.8661 &0.8811 \\
\textbf{Habkern,BE} &0.8688 &0.8749 &0.8652 &0.8783 \\
\textbf{Hägglingen,AG} &0.8744 &0.8804 &0.8708 &0.8893 \\
\textbf{Hallau,SH} &0.8732 &0.8780 &0.8683 &0.8885 \\
\end{tabular}
}
\end{table*}

\begin{table*}[t!]
\centering
\resizebox{0.8\textwidth}{!}{%
\begin{tabular}{l|rrrr|}\toprule
\multirow{2}{*}{\textbf{Swiss-German}} &\multicolumn{4}{c}{\textbf{COMET}} \\\cmidrule{2-5}
&\textbf{NLLB-Dis-600M} &\textbf{NLLB-Dis-1.3B} &\textbf{NLLB-1.3B} &\textbf{NLLB-3.3B} \\\midrule
\textbf{Lützelflüh,BE} &0.8654 &0.8705 &0.8631 &0.8807 \\
\textbf{Lungern,OW} &0.8672 &0.8733 &0.8645 &0.8799 \\
\textbf{Lupfig,AG} &0.8704 &0.8828 &0.8694 &0.8898 \\
\textbf{Thundorf,TG} &0.8742 &0.8909 &0.8751 &0.8928 \\
\textbf{Luzern,LU} &0.8712 &0.8772 &0.8684 &0.8851 \\
\textbf{Silenen,UR} &0.8740 &0.8800 &0.8667 &0.8873 \\
\end{tabular}
}
\end{table*}

\begin{table*}[t!]
\centering
\resizebox{0.75\textwidth}{!}{%
\begin{tabular}{l|rrrr|}\toprule
\multirow{2}{*}{\textbf{Swiss-German}} &\multicolumn{4}{c}{\textbf{COMET}} \\\cmidrule{2-5}
&\textbf{NLLB-Dis-600M} &\textbf{NLLB-Dis-1.3B} &\textbf{NLLB-1.3B} &\textbf{NLLB-3.3B} \\\midrule
\textbf{Pieterlen,BE} &0.8727 &0.8733 &0.8674 &0.8815 \\
\textbf{Plaffeien,FR} &0.8612 &0.8743 &0.8572 &0.8752 \\
\textbf{Pratteln,BL} &0.8666 &0.8728 &0.8651 &0.8839 \\
\textbf{Quarten,SG} &0.8757 &0.8870 &0.8758 &0.8921 \\
\textbf{Rafz,ZH} &0.8737 &0.8816 &0.8712 &0.8865 \\
\textbf{Ramsen,SH} &0.8748 &0.8809 &0.8724 &0.8866 \\
\end{tabular}
}
\end{table*}

\begin{table*}[t!]
\centering
\resizebox{0.75\textwidth}{!}{%
\begin{tabular}{l|rrrr|}\toprule
\multirow{2}{*}{\textbf{Swiss-German}} &\multicolumn{4}{c}{\textbf{COMET}} \\\cmidrule{2-5}
&\textbf{NLLB-Dis-600M} &\textbf{NLLB-Dis-1.3B} &\textbf{NLLB-1.3B} &\textbf{NLLB-3.3B} \\\midrule
\textbf{Sursee,LU} &0.8689 &0.8781 &0.8723 &0.8852 \\
\textbf{Täuffelen,BE} &0.8640 &0.8696 &0.8633 &0.8787 \\
\textbf{Tafers,FR} &0.8653 &0.8732 &0.8586 &0.8766 \\
\textbf{Tamins,GR} &0.8733 &0.8756 &0.8683 &0.8907 \\
\textbf{Teufenthal,AG} &0.8749 &0.8820 &0.8741 &0.8899 \\
\textbf{Thalwil,ZH} &0.8776 &0.8909 &0.8777 &0.8938 \\
\bottomrule
\end{tabular}
}
\caption{Compare COMET score of different Swiss-German dialects on a subset of 87 sentences.}
\label{tab:swiss-dialects-comp}
\end{table*}
\begin{table*}[t!]
\centering
\resizebox{0.9\textwidth}{!}{%
\begin{tabular}{l|r|rrrrr|}\toprule
\multirow{2}{*}{\textbf{Swiss-German}} &\multirow{2}{*}{\textbf{\# of Sentences}} &\multicolumn{4}{c}{\textbf{BLEU}} \\\cmidrule{3-6}
& &\textbf{NLLB-Dis-600M} &\textbf{NLLB-Dis-1.3B} &\textbf{NLLB-1.3B} &\textbf{NLLB-3.3B} \\\midrule
\textbf{Aarau,AG} &121 &42.68 &45.48 &41.85 &45.29 \\
\textbf{Aarberg,BE} &117 &43.83 &46.08 &41.73 &46.68 \\
\textbf{Aarburg,AG} &118 &43.51 &45.44 &42.02 &46.03 \\
\textbf{Adelboden,BE} &120 &41.16 &41.33 &39.97 &41.82 \\
\textbf{Aedermannsdorf,SO} &115 &43.34 &45.56 &41.56 &45.76 \\
\textbf{Aesch,BL} &118 &43.57 &44.50 &41.46 &45.56 \\
\end{tabular}
}
\end{table*}

\begin{table*}[t!]
\centering
\resizebox{0.9\textwidth}{!}{%
\begin{tabular}{l|r|rrrrr|}\toprule
\multirow{2}{*}{\textbf{Swiss-German}} &\multirow{2}{*}{\textbf{\# of Sentences}} &\multicolumn{4}{c}{\textbf{BLEU}} \\\cmidrule{3-6}
& &\textbf{NLLB-Dis-600M} &\textbf{NLLB-Dis-1.3B} &\textbf{NLLB-1.3B} &\textbf{NLLB-3.3B} \\\midrule
\textbf{Densbüren,AG} &121 &42.90 &45.41 &41.66 &45.67 \\
\textbf{Diemtigen,BE} &118 &43.25 &44.20 &42.60 &45.45 \\
\textbf{Diepoldsau,SG} &113 &44.76 &46.68 &42.86 &47.97 \\
\textbf{Diessbach bei Büren,BE} &115 &41.72 &44.78 &41.32 &45.89 \\
\textbf{Düdingen,FR} &114 &43.28 &43.40 &41.91 &46.62 \\
\textbf{Ebnat-Kappel,SG} &122 &44.41 &44.93 &42.36 &45.33 \\
\end{tabular}
}
\end{table*}

\begin{table*}[t!]
\centering
\resizebox{0.9\textwidth}{!}{%
\begin{tabular}{l|r|rrrrr|}\toprule
\multirow{2}{*}{\textbf{Swiss-German}} &\multirow{2}{*}{\textbf{\# of Sentences}} &\multicolumn{4}{c}{\textbf{BLEU}} \\\cmidrule{3-6}
& &\textbf{NLLB-Dis-600M} &\textbf{NLLB-Dis-1.3B} &\textbf{NLLB-1.3B} &\textbf{NLLB-3.3B} \\\midrule
\textbf{Schlatt-Haslen,AI} &112 &43.00 &45.35 &41.44 &46.68 \\
\textbf{Hedingen,ZH} &116 &43.58 &46.64 &42.48 &46.60 \\
\textbf{Heiden,AR} &118 &43.34 &46.14 &42.52 &46.02 \\
\textbf{Heitenried,FR} &118 &41.19 &43.01 &39.87 &44.37 \\
\textbf{Herisau,AR} &113 &44.67 &46.57 &42.95 &46.84 \\
\textbf{Hölstein,BL} &120 &43.77 &45.67 &41.34 &45.70 \\
\end{tabular}
}
\end{table*}

\begin{table*}[t!]
\centering
\resizebox{0.9\textwidth}{!}{%
\begin{tabular}{l|r|rrrrr|}\toprule
\multirow{2}{*}{\textbf{Swiss-German}} &\multirow{2}{*}{\textbf{\# of Sentences}} &\multicolumn{4}{c}{\textbf{BLEU}} \\\cmidrule{3-6}
& &\textbf{NLLB-Dis-600M} &\textbf{NLLB-Dis-1.3B} &\textbf{NLLB-1.3B} &\textbf{NLLB-3.3B} \\\midrule
\textbf{Magden,AG} &114 &42.59 &44.77 &41.13 &45.16 \\
\textbf{Maisprach,BL} &116 &43.95 &45.07 &42.59 &45.70 \\
\textbf{Malans,GR} &114 &43.78 &47.20 &42.79 &46.06 \\
\textbf{Malters,LU} &117 &42.62 &44.17 &40.39 &43.99 \\
\textbf{Mammern,TG} &120 &44.85 &47.15 &44.87 &47.44 \\
\textbf{Marbach,LU} &121 &44.63 &46.40 &43.94 &46.55 \\
\end{tabular}
}
\end{table*}

\begin{table*}[t!]
\centering
\resizebox{0.9\textwidth}{!}{%
\begin{tabular}{l|r|rrrrr|}\toprule
\multirow{2}{*}{\textbf{Swiss-German}} &\multirow{2}{*}{\textbf{\# of Sentences}} &\multicolumn{4}{c}{\textbf{BLEU}} \\\cmidrule{3-6}
& &\textbf{NLLB-Dis-600M} &\textbf{NLLB-Dis-1.3B} &\textbf{NLLB-1.3B} &\textbf{NLLB-3.3B} \\\midrule
\textbf{Reutigen,BE} &118 &43.08 &45.58 &42.86 &45.27 \\
\textbf{Rheineck,SG} &119 &43.50 &45.45 &42.15 &47.24 \\
\textbf{Medels im Rheinwald,GR} &111 &44.75 &47.27 &43.00 &46.92 \\
\textbf{Wattwil,SG} &117 &43.08 &46.42 &42.32 &46.12 \\
\textbf{Rickenbach,SO} &118 &42.66 &43.94 &41.46 &44.53 \\
\textbf{Rifferswil,ZH} &114 &43.75 &46.14 &43.29 &46.58 \\
\end{tabular}
}
\end{table*}

\begin{table*}[t!]
\centering
\resizebox{0.9\textwidth}{!}{%
\begin{tabular}{l|r|rrrrr|}\toprule
\multirow{2}{*}{\textbf{Swiss-German}} &\multirow{2}{*}{\textbf{\# of Sentences}} &\multicolumn{4}{c}{\textbf{BLEU}} \\\cmidrule{3-6}
& &\textbf{NLLB-Dis-600M} &\textbf{NLLB-Dis-1.3B} &\textbf{NLLB-1.3B} &\textbf{NLLB-3.3B} \\\midrule
\textbf{Triengen,LU} &118 &42.98 &44.26 &42.42 &44.36 \\
\textbf{Trimmis,GR} &117 &42.94 &45.24 &42.59 &47.24 \\
\textbf{Trogen,AR} &118 &43.28 &44.89 &41.74 &46.20 \\
\textbf{Tüscherz-Alfermée,BE} &115 &43.25 &45.20 &43.30 &45.97 \\
\textbf{Tuggen,SZ} &120 &45.74 &46.84 &43.51 &47.35 \\
\textbf{Turbenthal,ZH} &124 &44.83 &47.59 &44.28 &47.45 \\
\bottomrule
\end{tabular}
}
\caption{BLEU score of different Swiss-German dialects on all sentences.}
\label{tab:swiss-dialects_BLEU}
\end{table*}

\begin{table*}[t!]
\centering
\resizebox{0.8\textwidth}{!}{%
\begin{tabular}{l|rrrr|}\toprule
\multirow{2}{*}{\textbf{Swiss-German}} &\multicolumn{4}{c}{\textbf{BLEU}} \\\cmidrule{2-5}
&\textbf{NLLB-Dis-600M} &\textbf{NLLB-Dis-1.3B} &\textbf{NLLB-1.3B} &\textbf{NLLB-3.3B} \\\midrule
\textbf{Aarau,AG} &42.37 &44.92 &41.80 &45.08 \\
\textbf{Aarberg,BE} &43.92 &45.63 &41.64 &46.50 \\
\textbf{Aarburg,AG} &43.55 &45.04 &41.73 &45.80 \\
\textbf{Adelboden,BE} &41.32 &40.88 &40.20 &41.82 \\
\textbf{Aedermannsdorf,SO} &43.51 &45.18 &41.40 &45.76 \\
\textbf{Aesch,BL} &43.32 &44.28 &41.30 &45.70 \\
\end{tabular}
}
\end{table*}

\begin{table*}[t!]
\centering
\resizebox{0.8\textwidth}{!}{%
\begin{tabular}{l|rrrr|}\toprule
\multirow{2}{*}{\textbf{Swiss-German}} &\multicolumn{4}{c}{\textbf{BLEU}} \\\cmidrule{2-5}
&\textbf{NLLB-Dis-600M} &\textbf{NLLB-Dis-1.3B} &\textbf{NLLB-1.3B} &\textbf{NLLB-3.3B} \\\midrule
\textbf{Dagmersellen,LU} &43.37 &45.40 &41.86 &44.30 \\
\textbf{Davos,GR} &42.51 &48.28 &43.25 &47.52 \\
\textbf{Degersheim,SG} &43.95 &47.30 &43.46 &47.20 \\
\textbf{Densbüren,AG} &43.13 &45.45 &42.36 &46.12 \\
\textbf{Diemtigen,BE} &43.64 &44.05 &42.47 &45.65 \\
\textbf{Diepoldsau,SG} &44.82 &46.39 &43.17 &48.11 \\
\end{tabular}
}
\end{table*}

\begin{table*}[t!]
\centering
\resizebox{0.8\textwidth}{!}{%
\begin{tabular}{l|rrrr|}\toprule
\multirow{2}{*}{\textbf{Swiss-German}} &\multicolumn{4}{c}{\textbf{BLEU}} \\\cmidrule{2-5}
&\textbf{NLLB-Dis-600M} &\textbf{NLLB-Dis-1.3B} &\textbf{NLLB-1.3B} &\textbf{NLLB-3.3B} \\\midrule
\textbf{Gurtnellen,UR} &45.79 &47.38 &43.16 &47.10 \\
\textbf{Guttannen,BE} &41.24 &43.07 &44.10 &45.16 \\
\textbf{Guttet-Feschel,VS} &43.40 &43.76 &43.24 &45.38 \\
\textbf{Habkern,BE} &41.95 &43.22 &41.68 &43.27 \\
\textbf{Hägglingen,AG} &43.12 &44.81 &41.02 &44.39 \\
\textbf{Hallau,SH} &42.79 &43.39 &41.39 &45.57 \\
\end{tabular}
}
\end{table*}

\begin{table*}[t!]
\centering
\resizebox{0.8\textwidth}{!}{%
\begin{tabular}{l|rrrr|}\toprule
\multirow{2}{*}{\textbf{Swiss-German}} &\multicolumn{4}{c}{\textbf{BLEU}} \\\cmidrule{2-5}
&\textbf{NLLB-Dis-600M} &\textbf{NLLB-Dis-1.3B} &\textbf{NLLB-1.3B} &\textbf{NLLB-3.3B} \\\midrule
\textbf{Lützelflüh,BE} &41.08 &42.60 &40.63 &44.27 \\
\textbf{Lungern,OW} &41.99 &43.15 &40.55 &45.49 \\
\textbf{Lupfig,AG} &42.64 &46.24 &41.92 &46.65 \\
\textbf{Thundorf,TG} &44.29 &46.58 &43.28 &47.28 \\
\textbf{Luzern,LU} &43.36 &45.41 &42.75 &45.94 \\
\textbf{Silenen,UR} &44.72 &45.03 &41.89 &47.40 \\
\end{tabular}
}
\end{table*}

\begin{table*}[t!]
\centering
\resizebox{0.8\textwidth}{!}{%
\begin{tabular}{l|rrrr|}\toprule
\multirow{2}{*}{\textbf{Swiss-German}} &\multicolumn{4}{c}{\textbf{BLEU}} \\\cmidrule{2-5}
&\textbf{NLLB-Dis-600M} &\textbf{NLLB-Dis-1.3B} &\textbf{NLLB-1.3B} &\textbf{NLLB-3.3B} \\\midrule
\textbf{Pieterlen,BE} &43.94 &44.26 &41.64 &44.77 \\
\textbf{Plaffeien,FR} &40.10 &41.86 &39.09 &42.91 \\
\textbf{Pratteln,BL} &41.85 &43.90 &40.42 &45.64 \\
\textbf{Quarten,SG} &45.51 &46.65 &43.12 &48.35 \\
\textbf{Rafz,ZH} &43.35 &46.20 &43.41 &46.96 \\
\textbf{Ramsen,SH} &43.22 &43.17 &42.17 &44.32 \\
\end{tabular}
}
\end{table*}

\begin{table*}[t!]
\centering
\resizebox{0.8\textwidth}{!}{%
\begin{tabular}{l|rrrr|}\toprule
\multirow{2}{*}{\textbf{Swiss-German}} &\multicolumn{4}{c}{\textbf{BLEU}} \\\cmidrule{2-5}
&\textbf{NLLB-Dis-600M} &\textbf{NLLB-Dis-1.3B} &\textbf{NLLB-1.3B} &\textbf{NLLB-3.3B} \\\midrule
\textbf{Sursee,LU} &44.06 &45.50 &42.94 &46.22 \\
\textbf{Täuffelen,BE} &43.07 &43.73 &40.37 &44.37 \\
\textbf{Tafers,FR} &41.72 &42.16 &39.89 &43.73 \\
\textbf{Tamins,GR} &42.72 &44.48 &42.30 &47.25 \\
\textbf{Teufenthal,AG} &43.60 &44.72 &41.27 &44.33 \\
\textbf{Thalwil,ZH} &44.94 &48.49 &45.24 &47.56 \\
\bottomrule
\end{tabular}
}
\caption{Compare BLEU score of different Swiss-German dialects on a subset of 87 sentences.}
\label{tab:swiss-dialects-comp_BLEU}
\end{table*}
\begin{table*}[t!]
\centering
\resizebox{0.9\textwidth}{!}{%
\begin{tabular}{l|r|rrrrr|}\toprule
\multirow{2}{*}{\textbf{Swiss-German}} &\multirow{2}{*}{\textbf{\# of Sentences}} &\multicolumn{4}{c}{\textbf{COMET}} \\\cmidrule{3-6}
& &\textbf{NLLB-Dis-600M} &\textbf{NLLB-Dis-1.3B} &\textbf{NLLB-1.3B} &\textbf{NLLB-3.3B} \\\midrule
\textbf{AG} &3881 &0.8750 &0.8817 &0.8717 &0.8889 \\
\textbf{BE} &8389 &0.8691 &0.8758 &0.8665 &0.8853 \\
\textbf{SO} &1498 &0.8672 &0.8750 &0.8643 &0.8831 \\
\textbf{BL} &1867 &0.8703 &0.8740 &0.8657 &0.8840 \\
\textbf{VS} &2775 &0.8636 &0.8707 &0.8642 &0.8782 \\
\textbf{OW} &693 &0.8689 &0.8766 &0.8640 &0.8830 \\
\textbf{SZ} &1293 &0.8718 &0.8792 &0.8694 &0.8862 \\
\textbf{UR} &824 &0.8716 &0.8767 &0.8657 &0.8855 \\
\textbf{SG} &3522 &0.8726 &0.8819 &0.8714 &0.8870 \\
\textbf{TG} &2077 &0.8743 &0.8846 &0.8721 &0.8891 \\
\textbf{ZH} &4871 &0.8749 &0.8838 &0.8721 &0.8888 \\
\textbf{AI} &343 &0.8661 &0.8803 &0.8688 &0.8868 \\
\textbf{GR} &2677 &0.8733 &0.8800 &0.8697 &0.8875 \\
\textbf{BS} &228 &0.8719 &0.8832 &0.8676 &0.8893 \\
\textbf{SH} &1169 &0.8751 &0.8816 &0.8723 &0.8872 \\
\textbf{AR} &813 &0.8711 &0.8814 &0.8709 &0.8883 \\
\textbf{NW} &352 &0.8711 &0.8756 &0.8668 &0.8840 \\
\textbf{LU} &2565 &0.8714 &0.8773 &0.8689 &0.8869 \\
\textbf{FR} &1162 &0.8659 &0.8742 &0.8598 &0.8809 \\
\textbf{GL} &1091 &0.8761 &0.8839 &0.8733 &0.8924 \\
\textbf{ZG} &696 &0.8718 &0.8784 &0.8691 &0.8860 \\
\bottomrule
\end{tabular}
}
\caption{COMET score of different Swiss-German regions on all sentences.}
\label{tab:swiss-regions}
\end{table*}

\begin{table*}[t!]
\centering
\resizebox{0.8\textwidth}{!}{%
\begin{tabular}{l|rrrr|}\toprule
\multirow{2}{*}{\textbf{Swiss-German}} &\multicolumn{4}{c}{\textbf{COMET}} \\\cmidrule{2-5}
&\textbf{NLLB-Dis-600M} &\textbf{NLLB-Dis-1.3B} &\textbf{NLLB-1.3B} &\textbf{NLLB-3.3B} \\\midrule
\textbf{AG} &0.8742 &0.8820 &0.8720 &0.8887 \\
\textbf{BE} &0.8689 &0.8762 &0.8668 &0.8851 \\
\textbf{SO} &0.8666 &0.8751 &0.8640 &0.8827 \\
\textbf{BL} &0.8702 &0.8750 &0.8667 &0.8844 \\
\textbf{VS} &0.8637 &0.8715 &0.8647 &0.8790 \\
\textbf{OW} &0.8686 &0.8777 &0.8649 &0.8831 \\
\textbf{SZ} &0.8713 &0.8795 &0.8700 &0.8861 \\
\textbf{UR} &0.8711 &0.8771 &0.8662 &0.8852 \\
\textbf{SG} &0.8726 &0.8828 &0.8723 &0.8877 \\
\textbf{TG} &0.8743 &0.8853 &0.8732 &0.8896 \\
\textbf{ZH} &0.8747 &0.8844 &0.8728 &0.8892 \\
\textbf{AI} &0.8665 &0.8814 &0.8699 &0.8877 \\
\textbf{GR} &0.8729 &0.8801 &0.8700 &0.8874 \\
\textbf{BS} &0.8717 &0.8834 &0.8667 &0.8889 \\
\textbf{SH} &0.8747 &0.8819 &0.8731 &0.8872 \\
\textbf{AR} &0.8722 &0.8833 &0.8723 &0.8897 \\
\textbf{NW} &0.8712 &0.8768 &0.8682 &0.8842 \\
\textbf{LU} &0.8709 &0.8779 &0.8698 &0.8866 \\
\textbf{FR} &0.8656 &0.8748 &0.8609 &0.8808 \\
\textbf{GL} &0.8760 &0.8844 &0.8738 &0.8930 \\
\textbf{ZG} &0.8708 &0.8788 &0.8694 &0.8850 \\
\bottomrule
\end{tabular}
}
\caption{Comparable COMET score of different Swiss-German regions}
\label{tab:swiss-regions-comp}
\end{table*}

\begin{table*}[t!]
\centering
\resizebox{0.9\textwidth}{!}{%
\begin{tabular}{l|r|rrrrr|}\toprule
\multirow{2}{*}{\textbf{Swiss-German}} &\multirow{2}{*}{\textbf{\# of Sentences}} &\multicolumn{4}{c}{\textbf{BLEU}} \\\cmidrule{3-6}
& &\textbf{NLLB-Dis-600M} &\textbf{NLLB-Dis-1.3B} &\textbf{NLLB-1.3B} &\textbf{NLLB-3.3B} \\\midrule
\textbf{AG} &3881 &44.00 &46.28 &42.97 &46.16 \\
\textbf{BE} &8389 &42.79 &45.01 &41.97 &45.31 \\
\textbf{SO} &1498 &43.10 &45.58 &41.54 &45.63 \\
\textbf{BL} &1867 &43.42 &45.34 &41.52 &45.71 \\
\textbf{VS} &2775 &41.78 &43.27 &42.42 &44.77 \\
\textbf{OW} &693 &42.55 &44.55 &40.78 &45.95 \\
\textbf{SZ} &1293 &43.78 &46.06 &42.54 &46.53 \\
\textbf{UR} &824 &44.34 &45.54 &42.12 &46.90 \\
\textbf{SG} &3522 &43.94 &45.75 &42.71 &46.49 \\
\textbf{TG} &2077 &44.40 &46.66 &43.32 &46.56 \\
\textbf{ZH} &4871 &44.06 &46.87 &43.82 &46.82 \\
\textbf{AI} &343 &42.78 &44.66 &42.20 &47.14 \\
\textbf{GR} &2677 &43.79 &47.07 &43.46 &47.26 \\
\textbf{BS} &228 &43.33 &46.49 &43.34 &47.07 \\
\textbf{SH} &1169 &43.95 &45.26 &42.91 &45.84 \\
\textbf{AR} &813 &44.26 &45.51 &42.32 &46.56 \\
\textbf{NW} &352 &43.38 &44.47 &41.62 &45.33 \\
\textbf{LU} &2565 &43.25 &45.07 &41.95 &45.53 \\
\textbf{FR} &1162 &42.25 &43.47 &40.75 &45.20 \\
\textbf{GL} &1091 &44.22 &46.59 &43.60 &47.97 \\
\textbf{ZG} &696 &43.41 &45.95 &42.60 &46.72 \\
\bottomrule
\end{tabular}
}
\caption{BLEU score of different Swiss-German regions on all sentences.}
\label{tab:swiss-regions_BLEU}
\end{table*}

\begin{table*}[t!]
\centering
\resizebox{0.8\textwidth}{!}{%
%
}
\caption{Comparable BLEU score of different Swiss-German regions}
\label{tab:swiss-regions-comp_BLEU}
\end{table*}
\begin{table*}[t!]
\centering
\resizebox{\textwidth}{!}{
%
}
\caption{BLEU scores of different languages' dialects for various model scales.}
\label{tab:other-languages_BLEU}
\vspace{-1em}
\end{table*}

\end{document}